\documentclass{article}

 \usepackage[preprint]{neurips_2026}

\usepackage[utf8]{inputenc} % allow utf-8 input
\usepackage[T1]{fontenc}    % use 8-bit T1 fonts
\usepackage{url}            % simple URL typesetting
\usepackage{booktabs}       % professional-quality tables
\usepackage{amsfonts}       % blackboard math symbols
\usepackage{nicefrac}       % compact symbols for 1/2, etc.
\usepackage{microtype}      % microtypography
\usepackage[dvipsnames,table]{xcolor}        % colors

\definecolor{mycyan}{RGB}{212, 239, 251}
\usepackage[pagebackref=true,breaklinks=true,colorlinks,bookmarks=false, citecolor=Green]{hyperref}
\usepackage{multirow}
\usepackage{graphicx}
\usepackage{makecell}
\usepackage{amsmath}
\usepackage{wrapfig}
\usepackage{amssymb, mathtools}
\newtheorem{definition}{Definition}
\usepackage{pifont}
\newcommand{\cmark}{\ding{51}} % ✓
\newcommand{\xmark}{\ding{55}} % ✗
\usepackage{caption}

\title{\raisebox{-0.25\height}{\includegraphics[height=1.3em]{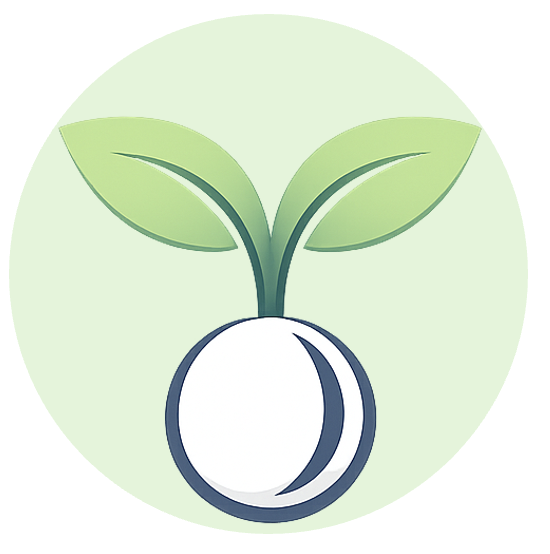}}\hspace{0.2em}SEED: Targeted Data Selection by Weighted Independent Set}

\author{
  \hspace{-0.25cm}Yuan Zhang$^{1}$, Lifeng Guo$^{2}$, Junwen Pan$^{3}$, Wenzhao Zheng$^{4}$, Wen Zhou$^{5}$ \\
  \hspace{-0.25cm}\textbf{Kuan Cheng$^{1}$, Kurt Keutzer$^{4}$, Shanghang Zhang$^{1}\thanks{Correspondence to: Shanghang Zhang.}$} \\
  \hspace{-0.25cm}$^1$ School of Computer Science, Peking University \\ $^2$ Beijing University of Posts and Telecommunications \quad $^3$ Tianjin University \\
  $^4$ EECS, UC Berkeley \quad $^5$ Chinese Academy of Sciences\\
  \url{https://github.com/Gumpest/SEED}
}

\begin{document}

\maketitle

\addtocontents{toc}{\protect\setcounter{tocdepth}{0}}
\begin{abstract}
Data selection seeks to identify a compact yet informative subset from large-scale training corpora, balancing sample quality against collection diversity.
% 引出WIS
We formulate this problem as a Weighted Independent Set (WIS) problem on a similarity graph, where nodes represent data samples weighted by influence, and edges connect semantically redundant pairs. This formulation naturally yields subsets that are simultaneously high-quality and diverse.
% 引出原始WIS的问题
However, two challenges arise in practice: naive node weights fail to distinguish informative signals from gradient noise, and edge construction under heterogeneous domain distributions produces structurally imbalanced graphs that bias selection toward sparse regions.
% 引出我们的方法
To address these issues, we introduce two principled refinements from a unified graph perspective: (1) \textit{node value calibration} that restricts influence estimation to the bilateral salient subspace to ground node importance in task-relevant signals rather than surface-level statistics; (2) \textit{local scale normalization} that adapts edge thresholds to local neighborhood density, mitigating graph imbalance induced by cross-domain distribution shifts.
Together, these components yield a robust and scalable data selection pipeline dubbed SEED.
% 数据集 + 实验
We further construct \texttt{Honeybee-Remake-SEED-200K}, a compact multimodal dataset curated by SEED. 
Extensive experiments show that SEED consistently outperforms state-of-the-art methods on instruction tuning, visual instruction tuning, and semantic segmentation across diverse model families.
\end{abstract}
\section{Introduction}
% 数据选择的动机
The rapid advancement of large-scale foundation models~\cite{dubey2024llama, kirillov2023segment, abdin2024phi, zhang2025sparsevlm, yang2025qwen3, wang2026multimodal} heavily relies on massive training corpora. However, naively scaling up training data incurs substantial computational overhead and often introduces redundant, noisy, or low-quality samples~\cite{albalak2024survey, liu2024less, engstrom2024dsdm, kang2024performance, zhang2025d3, feng2025tarot, chen2026neuron}. As model capacity increases, the marginal benefit of additional data may diminish rapidly, highlighting the need for effective data selection that improves efficiency in both training and storage.

% 引出WIS
Recent research has established a successful data selection strategy that delicately balances two fundamental objectives: individual sample quality \cite{xia2024less, matask, tandata} and overall collection diversity \cite{dai2025improving, zhang2025d3, maharana2023d2}. 
This trade-off can be naturally captured by a similarity graph, where nodes represent data samples weighted by their quality and edges are \textit{unweighted} connections between semantically redundant pairs, unlike prior graph-based methods~\cite{maharana2023d2} with weighted edges. Under this formulation, selecting a coreset reduces to finding a (Maximum) Weighted Independent Set (WIS) \cite{sanghavi2009message}.

\begin{figure}[t]
    \centering
    \vspace{1mm}
    \includegraphics[width=1.01\linewidth]{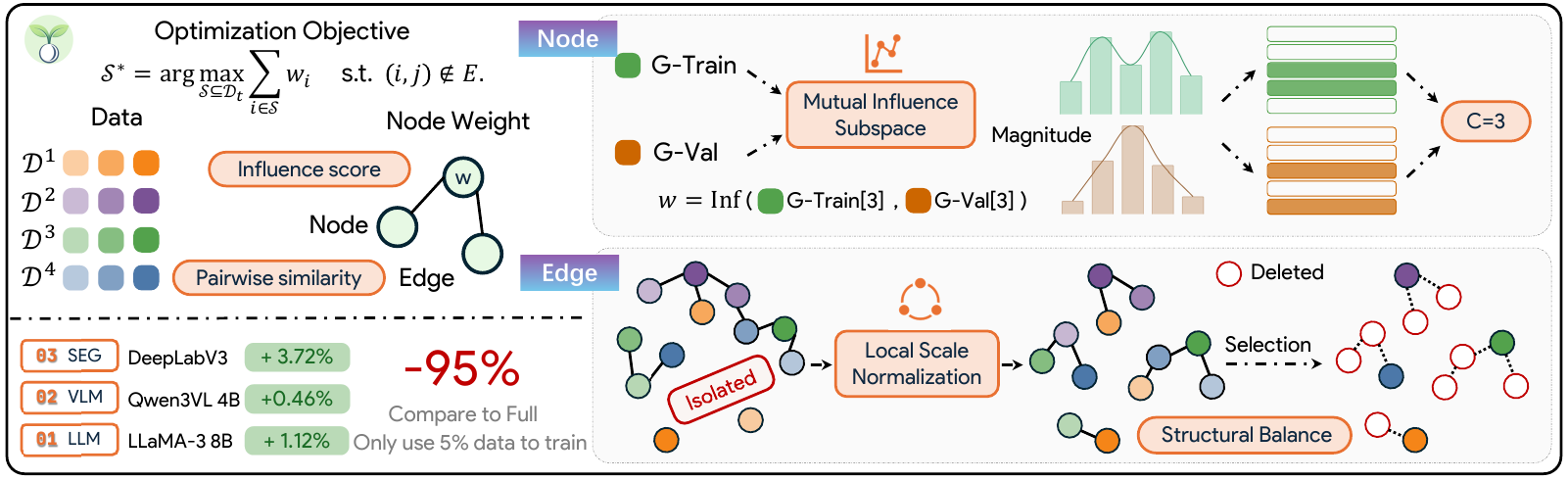}
    \caption{\textbf{Overview of SEED.} SEED formulates subset selection as a Weighted Independent Set problem over a similarity graph constructed from training data, with better node weights from a mutual influence subspace and better edges from local scale normalization. The resulting structurally balanced graph enables selecting a compact, diverse, and high-influence subset. Different colors indicate that nodes belong to different domains, while the color intensity represents the node weights.
    }
    \label{fig:pipeline}
    \vspace{0mm}
\end{figure}

% 直接使用WIS的问题
Despite its conceptual appeal, we identify two fundamental and previously undiagnosed failure modes for directly utilizing WIS.
\textbf{First, noisy and uncalibrated representations.}
Deep neural representations are inherently high-dimensional, yet gradient vectors exhibit a long-tail distribution across parameter dimensions~\cite{selvaraju2017grad, ismail2021improving} as shown in Figure~\ref{fig:mutual_influence_subspace}(a): a small subset of high-magnitude channels dominates the actual update direction, while the majority contribute little yet introduce noise into the influence. 
Furthermore, even a salient channel on the training side should be excluded if its counterpart magnitude on the validation side is negligible, such \textit{unilateral} dimensions distort rather than inform the influence estimate. 
Treating all channels uniformly thus leads to \textit{uncalibrated node representations} that corrupt true gradient influence and degrade node importance estimation.
\textbf{Second, structural imbalance in similarity graphs.}
As shown in Figure~\ref{fig:local_scale_motivation}(a), real-world datasets consist of multiple domains whose embeddings lie on distinct manifolds with highly non-uniform densities. This heterogeneity is further quantified in Figure~\ref{fig:local_scale_motivation}(b), where the distribution of $k$-th neighbor distances varies significantly across domains, indicating inconsistent local scales. Constructing edges via a global similarity threshold in such settings results in imbalanced graph structures: dense regions collapse into highly connected subgraphs, while sparse regions remain weakly connected or isolated. This imbalance biases the independent set selection, \textit{suppressing representative samples in dense regions while over-selecting isolated outliers}.

To overcome these limitations, we introduce SEED, a robust and scalable data \underline{SE}lection pipeline that refines the W\underline{E}ighted In\underline{D}ependent Set formulation from a unified graph perspective, incorporating two principled enhancements.
(1) \textbf{Nodes}: we propose node value calibration\footnote{In our paper, the influence score \cite{pruthi2020estimating} is used for node value.}, which restricts influence estimation to the bilateral salient subspace shared by training and validation gradients. Concretely, a channel is retained only if its gradient magnitude is consistently salient on \emph{both} splits, filtering out unilateral and noise-dominated dimensions. This mutual subspace restriction yields more reliable node importance estimates and effectively addresses challenge~1. (2) \textbf{Edges}: we introduce local scale normalization, which adjusts edge construction according to the local neighborhood distribution of each node. Instead of using a global similarity threshold, we normalize distances via local scaling to mitigate the graph imbalance induced by cross-domain distribution shifts, ensuring fair representation across both dense and sparse semantic regions, as shown in Figure~\ref{fig:local_scale_motivation}(c). With these two refinements, our resulting method, SEED, as illustrated in Figure~\ref{fig:pipeline}, is general, effective, and can be readily applied across a wide range of practical scenarios.

Extensive experiments show SEED consistently surpasses current state-of-the-art data selection (or pruning) methods on standard model training settings for instruction tuning, visual instruction tuning, and semantic segmentation tasks. For instance, SEED obtains $41.9$ EM with LLaMA-2-7B~\cite{touvron2023llama} on TyDiQA~\cite{tydiqa} under a $5\%$ selection ratio, surpassing InfoMax~\cite{tandata} by $2.3$ points; while on semantic segmentation, selecting only $20\%$ of the data using SEED achieves a $3.01$ mIoU improvement over full-data training with DeepLabV3~\cite{chen2017rethinking} on Cityscapes \cite{cordts2016cityscapes}. Moreover, to demonstrate practical usage in training, we implement SEED in a small proxy setting. For example, at matching accuracy levels, SEED requires only $2.5\%$ of the training data with the Qwen3-1.7B \cite{yang2025qwen3} proxy model and the Phi-4-14B \cite{abdin2024phi} target model on instruction tuning tasks, reducing GPU costs by $\mathbf{2.5\times}$.
% 数据集贡献
% As a further contribution, building upon Honeybee-1M~\cite{bansal2025honeybee}, we construct a large-scale multimodal dataset for visual instruction tuning with ground-truth annotations refreshed using Doubao-1.6-VL~\cite{guo2025seed1}, and apply SEED to vote for the top $20\%$, yielding \texttt{Honeybee-Remake-SEED-200K}.
As a further contribution, we apply SEED to Honeybee-1M~\cite{bansal2025honeybee} and curate \texttt{Honeybee-Remake-SEED-200K}, a compact multimodal dataset for visual instruction tuning that retains only $20\%$ of the original data yet improves performance across multiple multimodal evaluation benchmarks.

\section{Preliminaries}
\label{sec:prelims}

\subsection{Trajectory-based Data Influence}
\label{sec:inf}
To quantify the contribution of a training sample to a target task, we adopt a first-order approximation of training dynamics following \cite{pruthi2020estimating}, which avoids the expensive Hessian-vector products required by classical influence functions \cite{koh2017understanding}.

\noindent\textbf{Per-step Influence.}
Consider a model with parameters $\theta \in \mathbb{R}^d$ trained via stochastic gradient descent (SGD). At each step $t$, the parameters are updated as
$\theta_{t+1} = \theta_t - \eta_t \nabla_{\theta} \mathcal{L}(z^{(t)}; \theta_t)$, where $z^{(t)}$ is the training sample used at step $t$ and $\eta_t$ is the learning rate. To measure the effect of this update on a target sample $z'$, we apply a first-order Taylor expansion of $\mathcal{L}(z'; \theta)$ around $\theta_t$:
\begin{equation}
    \mathcal{L}(z'; \theta_{t+1}) - \mathcal{L}(z'; \theta_t)
    \approx
    \langle \nabla_{\theta} \mathcal{L}(z'; \theta_t),\,
            \theta_{t+1} - \theta_t \rangle.
\end{equation}
When substituting the SGD update rule, the loss change on $z'$ induced by training on $z^{(t)}$ is $-\eta_t \langle \nabla_{\theta}\mathcal{L}(z^{(t)};\theta_t),\, \nabla_{\theta}\mathcal{L}(z';\theta_t)\rangle$. We define the per-step influence as the \emph{reduction} in this loss, \emph{i.e.}, the negative of the loss change:
\begin{equation}
    \mathcal{I}_{\text{step}}(z^{(t)}, z')
    \coloneqq
    \eta_t \langle \nabla_{\theta} \mathcal{L}(z^{(t)}; \theta_t),\,
                   \nabla_{\theta} \mathcal{L}(z'; \theta_t) \rangle.
\end{equation}
A positive value indicates that the gradients of $z^{(t)}$ and $z'$ are aligned, meaning the update on $z^{(t)}$ decreases the loss on $z'$; a negative value indicates the opposite.

\noindent\textbf{Trajectory-level Influence.}
Let $\mathcal{T}(z)$ denote the set of steps at which sample $z$ is used during training. The overall influence of $z$ on $z'$ is obtained by accumulating per-step contributions across the training trajectory:
\begin{equation}
    \text{Inf}_{\text{traj}}(z, z')
    \coloneqq
    \sum_{t \in \mathcal{T}(z)}
    \eta_t \langle \nabla_{\theta} \mathcal{L}(z; \theta_t),\,
                   \nabla_{\theta} \mathcal{L}(z'; \theta_t) \rangle.
    \label{eq:total_inf}
\end{equation}
Since computing Eq.~\ref{eq:total_inf} over all training steps is expensive, we approximate it using a sparse set of checkpoints sampled from the trajectory \cite{pruthi2020estimating}, balancing computational cost and estimation fidelity.

\subsection{Weighted Independent Set}
\label{sec:prelim_mwis}
The (Maximum) Weighted Independent Set (WIS) problem is a classical combinatorial optimization task defined on an undirected graph $G = (V, E)$ with non-negative node weights $\{w_v\}_{v \in V}$.

\noindent\textbf{Problem Formulation.}
An \emph{independent set} is a subset of vertices $S \subseteq V$ such that no two vertices in $S$ are adjacent. The WIS problem seeks an independent set that maximizes the total weight:
\begin{equation}
S^* = \arg\max_{S \subseteq V} \sum_{v \in S} w_v
\quad
\text{s.t.} \quad
(u, v) \notin E, \ \forall u, v \in S.
\label{eq:mwis_obj}
\end{equation}
Therefore, WIS naturally balances node importance (weights) and pairwise conflicts (edges).

\noindent\textbf{Efficient Implementation.}
While finding the exact $S^*$ is NP-hard for general graphs \cite{garey1979computers}, we adopt an efficient \textit{greedy approximation} algorithm \cite{halldorsson1994greed} powered by GPU-accelerated similarity search.
Briefly, we construct a sparse conflict graph by computing pairwise influence using FAISS \cite{johnson2019billion}, followed by a max-heap-based greedy selection. Details can be found in Appendix \ref{sec:app_mwis_impl}.
\section{Method}
\label{sec:method}

\subsection{Weighted Independent Set among Datasets}
\label{sec:wis}

Given a training dataset $\mathcal{D}_{\text{train}} = \{z_i\}_{i=1}^{N}$ and a target dataset $\mathcal{D}_{\text{target}} = \{z'_j\}_{j=1}^{M}$, our goal is to select a compact subset $\mathcal{S} \subseteq \mathcal{D}_{\text{train}}$ that is both informative for the target task and internally diverse. We achieve this by incorporating trajectory-based influence (§\ref{sec:inf}) into the WIS framework (§\ref{sec:prelim_mwis}).

\begin{definition}[Node Weights]
    \label{def:node_weights}
    The weight $w_v$ of a node $v \in \mathcal{D}_{\text{train}}$ is defined as its influence on the target dataset $\mathcal{D}_{\text{target}}$.
\end{definition}

Following Definition~\ref{def:node_weights}, we quantify the importance of each training sample $z_i$ via its aggregated influence on the target dataset. Specifically, we set the node weight $w_i$ to the maximum trajectory-level influence across all target samples:
\begin{equation}
    w_i \coloneqq \max_{z' \in \mathcal{D}_{\text{target}}} 
    \operatorname{Inf}_{\text{traj}}(z_i, z').
    \label{eq:node_weight}
\end{equation}
Intuitively, a training sample $z_i$ receives a high node weight if it substantially benefits at least one target sample, thereby reflecting its relevance to the target task.

\begin{definition}[Edges]
    \label{def:edge}
    An edge $(u, v)$ connects two distinct nodes $u, v \in \mathcal{D}_{\text{train}}$ if their mutual influence is positive, suggesting potential redundancy.
\end{definition}

To capture redundancy between training samples, we construct a conflict graph $G = (\mathcal{D}_{\text{train}}, E)$ where an edge $(i, j) \in E$ is added if the mutual influence between the two samples is positive:
\begin{equation}
    (i, j) \in E \iff \operatorname{Inf}_{\text{train}}(z_i, z_j) - \tau > 0,
    \label{eq:edge}
\end{equation}
where $\operatorname{Inf}_{\text{train}}(\cdot, \cdot)$ quantifies the pairwise influence between training samples, and $\tau$ is a predefined threshold controlling the sparsity of the conflict graph. To scale edge construction to large datasets, we restrict comparisons to the $k$ nearest neighbors of each node, yielding a sparse conflict graph.

Finally, given node weights $\{w_v\}$ and an unweighted conflict graph $G = (\mathcal{D}_{\text{train}}, E)$, we formulate subset data selection as:
\begin{equation}
    \mathcal{S}^* = \arg\max_{\substack{\mathcal{S} \subseteq \mathcal{D}_{\text{train}} \\ |\mathcal{S}| \leq K}} 
    \sum_{z_i \in \mathcal{S}} \left( \max_{z' \in \mathcal{D}_{\text{target}}} \operatorname{Inf}_{\text{traj}}(z_i, z') \right)
    \text{s.t.} \quad 
    \operatorname{Inf}_{\text{train}}(z_u, z_v) \le \tau,\ 
    \forall u \neq v,\ z_u, z_v \in \mathcal{S},
    \label{eq:wis_inf_formulation}
\end{equation}
where $K$ is the target selection budget. This objective simultaneously maximizes the task-relevant influence of the selected subset while enforcing diversity by prohibiting mutually redundant pairs.

\subsection{Node Calibration: Mutual Subspace Restriction}
\label{sec:node_calibration}
\begin{figure}
    \centering
    \vspace{0mm}
    \includegraphics[width=1\linewidth]{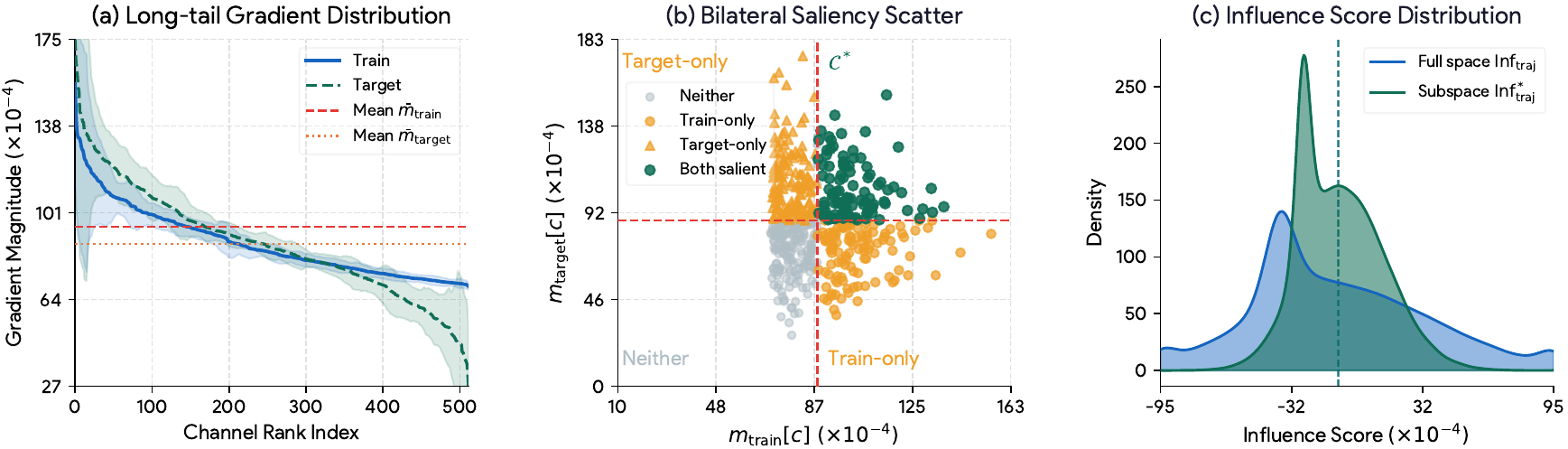}
    \caption{
    \textbf{Mutual Influence Subspace.}
    (a) Gradient magnitudes across channels show a long-tail distribution. 
    (b) Mutual space on both training and target (1st quadrant) form \(\mathcal{C}^*\). 
    (c) KDE of influence scores shows that restricting to \(\mathcal{C}^*\) reduces noise and sharpens the distribution.
    }
    \label{fig:mutual_influence_subspace}
    \vspace{-1mm}
\end{figure}

Standard trajectory-level influence adopts gradient inner products over the full parameter space, implicitly treating all dimensions as equally informative. As shown in Figure~\ref{fig:mutual_influence_subspace}(a), gradient vectors exhibit a long-tail distribution across parameter dimensions: a small subset of high-magnitude channels dominates the actual update direction, while the majority of low-magnitude channels contribute little to updates yet introduce noise into the inner product, distorting influence estimation.

To obtain more accurate node weights, we therefore design a subspace to mitigate this interference, computing the influence of the training set on the target dataset strictly within this subspace.

Formally, given the gradient matrix of the training set $\mathcal{G}_{\text{train}} \in \mathbb{R}^{N \times C}$, we estimate the 
per-channel saliency by computing the mean absolute gradient magnitude across nodes:
\begin{equation}
    m_{\text{train}}[c] \coloneqq \frac{1}{N} \sum_{i=1}^{N} |\mathcal{G}_{\text{train}}[i, c]|.
    \label{eq:channel_saliency}
\end{equation}

However, even if a channel exhibits a large magnitude on the training sample $z$, if its corresponding magnitude is negligible on the target sample $z'$, its actual influence on $z'$ is minimal in this space. In this case, the contribution of the channel in $z$ is \textit{unilateral} and should not be counted in the alignment measure, as illustrated by the points in the 2nd–4th quadrants of Figure~\ref{fig:mutual_influence_subspace}(b).

Therefore, a channel space $c$ is retained only if it exceeds the mean saliency on \emph{both} splits, ensuring that the selected dimensions simultaneously drive updates on the training and target sides. Given the target set $\mathcal{G}_{\text{target}} \in \mathbb{R}^{M \times C}$, the resulting mutual subspace is defined as:
\begin{equation}
    \mathcal{C}^* \coloneqq 
    \left\{ c \;\middle|\; 
        m_{\text{train}}[c] > \bar{m}_{\text{train}} 
        \text{ , }\;
        m_{\text{target}}[c] > \bar{m}_{\text{target}}
    \right\},
    \label{eq:channel_mask}
\end{equation}
where $m_{\text{target}}[c] \coloneqq \frac{1}{M} \sum_{j=1}^{M} |\mathcal{G}_{\text{target}}[j, c]|$ denotes the mean saliency of channel $c$ over the target set.

Eventually, the influence score $\text{Inf}_{\text{traj}}^*(z_i, z'_j)$ is recomputed on the selected mutual subspace $\mathcal{C}^*$:
\begin{equation}
    \text{Inf}_{\text{traj}}^*(z_i, z'_j) \coloneqq 
    \sum_{t \in \mathcal{T}(z_i)} \eta_t \,
    \langle \nabla_{\mathcal{C}^*} \mathcal{L}(z_i; \theta_t),\,
            \nabla_{\mathcal{C}^*} \mathcal{L}(z'_j; \theta_t) \rangle,
    \label{eq:calibrated_inf}
\end{equation}
where $\nabla_{\mathcal{C}^*}$ denotes the gradient restricted to channels in $\mathcal{C}^*$. By projecting onto the bilateral salient subspace (the 1st quadrants of Figure~\ref{fig:mutual_influence_subspace}(b)), we suppress noise-dominated dimensions and emphasize shared update directions, obtaining more reliable influence estimates for nodes in WIS framework. 

The effect of this restriction is visualized in Figure~\ref{fig:mutual_influence_subspace}(c), which shows the kernel density estimation (KDE) \cite{kim2012robust} of influence scores in the full parameter space versus the bilateral salient subspace. Restricting to $\mathcal{C}^*$ clearly suppresses extreme noise and sharpens the distribution, highlighting that the selected subspace captures the most meaningful contributions for node influence estimation.

\subsection{Edge Construction: Local Scale Normalization}
% \subsection{Local Scale Normalization}
\label{sec:Edge_Construction}
\begin{figure}[t]
    \centering
    \vspace{0mm}
    \includegraphics[width=1\linewidth]{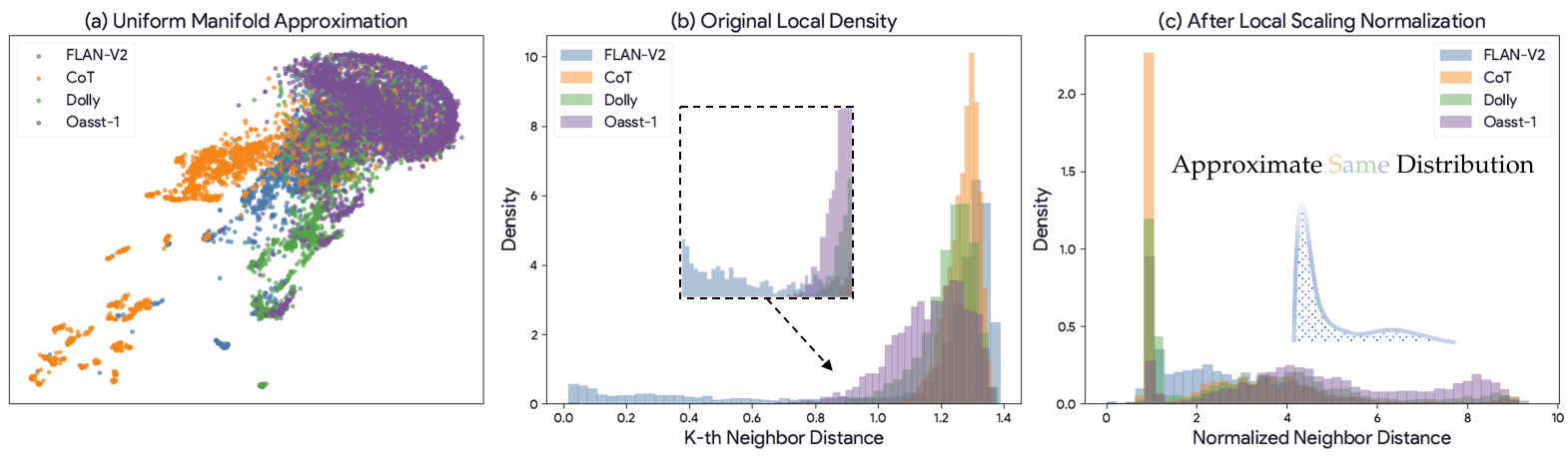}
    \caption{
        \textbf{Visualization of density heterogeneity across domains.} The data are sampled from FLAN-V2 \cite{longpre2023flan}, CoT \cite{wei2022chain}, Dolly \cite{DatabricksBlog2023DollyV2}, and Oasst-1 \cite{kopf2023openassistant}. (a) Embeddings from different domains exhibit non-uniform densities. (b) Global distance statistics vary significantly across domains. (c) Local scaling mitigates this mismatch by aligning neighborhood distributions. Best viewed in color.
    }
    \label{fig:local_scale_motivation}
    \vspace{-1mm}
\end{figure}

As illustrated in Figure~\ref{fig:local_scale_motivation}, real-world training corpora consist of multiple domains with highly non-uniform embedding densities. Applying a global predefined threshold $\tau$ of MWIS to such data yields structurally imbalanced conflict graphs: dense domains (\emph{e.g.}, Oasst-1 \cite{kopf2023openassistant}) accumulate excessive edges and become over-suppressed, while sparse domains (\emph{e.g.}, CoT \cite{wei2022chain}) remain under-connected and tend to be over-selected, degrading both diversity and downstream performance.

Therefore, we propose local-scale normalization to address this, which dynamically adapts the edge criterion to the local neighborhood structure of each node in MWIS. For each training sample $z_i$, we first estimate its local density $\sigma_i$ by measuring its influence on the $k$-th nearest neighbor:
\begin{equation}
    \sigma_i \coloneqq \operatorname{Inf}_{\text{train}}(z_i, z_i^{(k)}),
    \label{eq:sigma}
\end{equation}
We assume that larger influence values correspond to stronger similarity and, consequently, higher local density. Here, $z_i^{(k)}$ denotes the $k$-th nearest neighbor of $z_i$ in $\mathcal{D}_{\text{train}}$. Accordingly, a large $\sigma_i$ indicates that $z_i$ lies in a dense region, while a small $\sigma_i$ suggests a sparse neighborhood.

Furthermore, rather than applying a fixed threshold $\tau$ globally, we define a per-node level adaptive threshold that relaxes the edge formation criterion in dense regions:
\begin{equation}
    \tau_i \coloneqq \max(\tau,\; \alpha \cdot \sigma_i),
    \label{eq:adaptive_threshold}
\end{equation}
where $\alpha \in (0, 1)$ is a constant scaling factor. An edge $(i, j)$ is then added to the conflict graph only when 
\begin{equation}
    (i, j) \in E \iff 
\operatorname{Inf}_{\text{train}}(z_i, z_j) - \max(\tau_i, \tau_j) > 0
\end{equation}

By increasing $\tau_i$ in dense regions and reverting to $\tau$ in sparse regions, we obtain a density-aware edge construction strategy that suppresses redundant influences while preserving global connectivity, thereby improving the trade-off between influence maximization and diversity in the MWIS objective.

\subsection{Weighted Independent Set Formulation: SEED}
\label{sec:refined_mwis}

Integrating the mutual influence subspace and local scale normalization into the WIS framework, we obtain a refined and practical formulation for subset selection, as illustrated in Figure~\ref{fig:pipeline}.

Building upon §\ref{sec:wis}, we redefine the node weight $w_i$ for each training sample $z_i$ using the subspace-restricted influence $\operatorname{Inf}_{\text{traj}}^*(\cdot, \cdot)$ in Eq.~\eqref{eq:calibrated_inf}:
\begin{equation}
    w_i^* \coloneqq \max_{z' \in \mathcal{D}_{\text{target}}} \operatorname{Inf}_{\text{traj}}^*(z_i, z').
    \label{eq:refined_node_weight}
\end{equation}
To construct the conflict graph $G^* = (\mathcal{D}_{\text{train}}, E^*)$, an edge between $z_i$ and $z_j$ is established adaptively based on the local density condition described in Eq.~\eqref{eq:adaptive_threshold}:
\begin{equation}
    (i, j) \in E^* \iff \operatorname{Inf}_{\text{train}}(z_i, z_j) > \max(\tau_i, \tau_j).
    \label{eq:refined_edge}
\end{equation}
The final optimization problem is then formulated as a WIS over the refined graph $G^*$:
\begin{equation}
    \mathcal{S}^* = \arg\max_{\substack{\mathcal{S} \subseteq \mathcal{D}_{\text{train}} \\ |\mathcal{S}| \leq K}} 
    \sum_{z_i \in \mathcal{S}} w_i^*
    \quad \text{s.t.} \quad \forall z_u, z_v \in \mathcal{S},\; (u, v) \notin E^*.
    \label{eq:refined_wis_formulation}
\end{equation}
We term the resulting framework \textbf{SEED}, which bridges the WIS-based formulation and principled subset selection via subspace-calibrated node weighting and density-adaptive edge modeling.
\section{Experiments}
\label{sec:exp}
To comprehensively validate the generalization of our SEED, we conduct extensive experiments on instruction tuning, visual instruction tuning, and semantic segmentation tasks. Each reported result is averaged over \textbf{five} independent runs, with a standard deviation below $0.85$.

\subsection{SEED for Instruction Tuning}
\textbf{Settings.}
Following the experimental setup of LESS \cite{xia2024less}, we conduct data selection from FLAN-V2 \cite{longpre2023flan}, CoT \cite{wei2022chain}, Dolly \cite{DatabricksBlog2023DollyV2}, and OpenAssistant-1 \cite{kopf2023openassistant}. TyDiQA \cite{tydiqa}, MMLU \cite{hendrycks2020measuring}, and BBH \cite{suzgun2023challenging} are used as the target evaluation tasks. LLaMA-2-7B \cite{touvron2023llama}, LLaMA-3-8B \cite{dubey2024llama}, and Qwen3-4B \cite{yang2025qwen3} are the base models for both $5\%$ and $1\%$ data selection and fine-tuning. We employ Exact Match (EM) for TyDiQA and BBH, and Accuracy for MMLU.

% \textbf{Evaluation Metrics.} We employ Exact Matching (EM) for
% TyDiQA, BBH, and Accuracy for MMLU.

\definecolor{mygray}{gray}{.95}
\definecolor{prunegreen}{RGB}{0,120,60}
\vspace{-0.5mm}
\begin{table*}[ht!]
\caption{\textbf{Performance of SEED on instruction tuning for LLMs.} Best results are in \textbf{bold}.}
\vspace{-1.5mm}
\label{tab:res_llm}
\centering
\renewcommand{\arraystretch}{1.2} % Adjust row spacing
\setlength{\tabcolsep}{1.1mm}
\resizebox{\textwidth}{!}{
\begin{tabular}{p{2.6cm}p{.75cm}|ccc|ccc|ccc|c}
    \toprule[0.95pt]
    \multicolumn{2}{c|}{\multirow{2}{*}[-1.0ex]{\textbf{Method}}}
    &
    \multicolumn{3}{c|}{\raisebox{-0.25\height}{\includegraphics[height=1.05em]{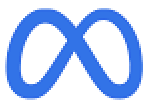}}\hspace{0.2em}\scshape\textbf{LLaMA-2-7B}}
    & \multicolumn{3}{c|}{\raisebox{-0.25\height}{\includegraphics[height=1.05em]{fig/llama.png}}\hspace{0.2em}\scshape\textbf{LLaMA-3-8B}}
    & \multicolumn{3}{c|}{\raisebox{-0.25\height}{\includegraphics[height=1.3em]{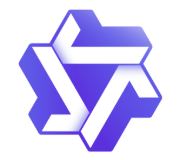}}\hspace{0.2em}\scshape\textbf{Qwen3-4B}}
    & \multirow{2}{*}[-1.0ex]{\scshape\textbf{Avg.}} \\ 
    \cmidrule(lr){3-5}
    \cmidrule(lr){6-8}
    \cmidrule(lr){9-11}
    & &
    \scshape\textbf{TyDiQA}
    & \scshape\textbf{MMLU}
    & \scshape\textbf{BBH}
    & \scshape\textbf{TyDiQA}
    & \scshape\textbf{MMLU}
    & \scshape\textbf{BBH}
    & \scshape\textbf{TyDiQA}
    & \scshape\textbf{MMLU}
    & \scshape\textbf{BBH}
    & \\
    \cmidrule(lr){1-2}
    \cmidrule(lr){3-5}
    \cmidrule(lr){6-8}
    \cmidrule(lr){9-11}
    \cmidrule(lr){12-12}
    \scshape\textbf{Base} & \texttt{\scriptsize{(0\%)}} & 32.8 & 45.7 & 39.5 & 47.3 & 63.3 & 62.2 & 47.4 & 67.9 & 70.0 & 53.0 \\
    \rowcolor{mygray}
    \scshape\textbf{Full} & \texttt{\scriptsize{(100\%)}} & 40.5 & 50.4 & 44.8 & 55.1 & 65.1 & 67.3 & 49.5 & 70.0 & 72.9 & 57.3 \\
    \midrule[0.6pt]
    \multicolumn{12}{c}{{\textbf{\large {\color{prunegreen}5\%}\,\textsc{Selection}}}}\\
    \scshape\textbf{Random} & & 38.5 & 46.8 & 40.0 & 51.6 & 64.0 & 63.1 & 48.5 & 68.7 & 70.3 & 54.6 \\
    \scshape\textbf{$\text{D}^2$Pruning} \cite{maharana2023d2} & \texttt{\scriptsize{ICLR24}} 
    & 39.0 & 48.9 & 40.3 & 54.8 & 64.2 & 65.0 & 49.3 & 68.5 & 70.4 & 55.6 \\
    \scshape\textbf{LESS} \cite{xia2024less} & \texttt{\scriptsize{ICML24}} & 39.3 & 49.3 & 40.6 & 55.4 & 64.4 & 65.7 & 49.6 & 68.8 & 70.6 & 55.9 \\
    \scshape\textbf{InfoMax} \cite{tandata} & \texttt{\scriptsize{ICLR25}} & 39.6 & \textbf{50.1} & 41.5 & 55.7 & 64.5 & 66.1 & 49.9 & 69.1 & 71.3 & 56.4 \\
    \scshape\textbf{TAROT} \cite{feng2025tarot} & \texttt{\scriptsize{ICML25}} & 39.7 & 49.5 & 40.9 & 55.2 & 64.7 & 66.0 & 49.6 & 69.0 & 71.0 & 56.2 \\
    \scshape\textbf{Nait} \cite{chen2026neuron} & \texttt{\scriptsize{ICLR26}} & 39.9 & 49.0 & 41.3 & 55.9 & 64.4 & 66.1 & 50.2 & 68.5 & 70.8 & 56.2 \\
    \rowcolor{mygray}
    \scshape\textbf{SEED} & & \textbf{41.9} & 49.6 & \textbf{41.9} & \textbf{57.6} & \textbf{65.1} & \textbf{66.9} & \textbf{50.6} & \textbf{69.5} & \textbf{72.6} & \textbf{57.3} \\
    \midrule[0.6pt]
    \multicolumn{12}{c}{{\textbf{\large {\color{prunegreen}1\%}\,\textsc{Selection}}}}\\
    \scshape\textbf{Random} & & 39.6 & 46.1 & 39.8 & 52.3 & 65.3 & 66.3 & 52.0 & 68.7 & 73.9 & 56.0 \\
    \scshape\textbf{$\text{D}^2$Pruning} \cite{maharana2023d2} & \texttt{\scriptsize{ICLR24}}  & 40.0 & 46.3 & 40.6 & 54.7 & 65.5 & 67.3 & 52.3 & 68.9 & 74.0 & 56.6 \\
    \scshape\textbf{LESS} \cite{xia2024less} & \texttt{\scriptsize{ICML24}} & 40.3 & 45.8 & 40.1 & 54.1 & 65.6 & 67.8 & 52.6 & 69.0 & 73.8 & 56.5 \\
    \scshape\textbf{InfoMax} \cite{tandata} & \texttt{\scriptsize{ICLR25}} & 40.8 & 46.5 & \textbf{41.5} & 54.6 & 65.8 & 68.2 & 53.1 & 69.3 & 74.2 & 57.1 \\
    \scshape\textbf{MoNA} \cite{matask} & \texttt{\scriptsize{NIPS25}} & 40.9 & \textbf{47.0} & 41.0 & 55.0 & 65.4 & 68.1 & \textbf{54.4} & 69.0 & 74.3 & 57.2 \\
    \scshape\textbf{Nait} \cite{chen2026neuron} & \texttt{\scriptsize{ICLR26}} & 40.1 & 46.2 & 40.7 & 55.4 & 65.1 & 67.4 & 53.3 & 69.2 & 73.6 & 56.8 \\
    \rowcolor{mygray}
    \scshape\textbf{SEED} & & \textbf{42.5} & 46.9 & 41.4 & \textbf{56.1} & \textbf{66.6} & \textbf{69.9} & \textbf{54.4} & \textbf{70.4} & \textbf{75.9} & \textbf{58.2} \\
    \bottomrule[0.95pt]
\end{tabular}
}
\vspace{-0.5mm}
\end{table*}

\textbf{Main Results.}
Table~\ref{tab:res_llm} shows that SEED achieves the best average performance under both $5\%$ and $1\%$ selection settings. At $5\%$, SEED attains an average of $57.3$, matching full-data fine-tuning and surpassing all baselines, including TAROT~\cite{feng2025tarot} and Nait~\cite{chen2026neuron} by $1.1$ points. Under the stricter $1\%$ setting, SEED achieves $58.2$, exceeding the strongest competitor MoNA~\cite{matask} by $1.0$ points, with particularly strong gains on BBH (\textit{e.g.}, $75.9$ on Qwen3-4B, surpassing InfoMax~\cite{tandata} by $1.7$ points).

Notably, performance at $5\%$ selection is slightly lower than at $1\%$. One possible explanation is that enlarging the subset may incorporate noisier or less task-aligned instances from heterogeneous sources, which weakens the advantage of carefully curated samples. This interesting finding suggests that the selection ratio itself plays a pivotal role in instruction tuning and warrants deeper investigation.

\begin{figure}[t]
    \centering
    \vspace{-1.5mm}
    \includegraphics[width=0.9\linewidth]{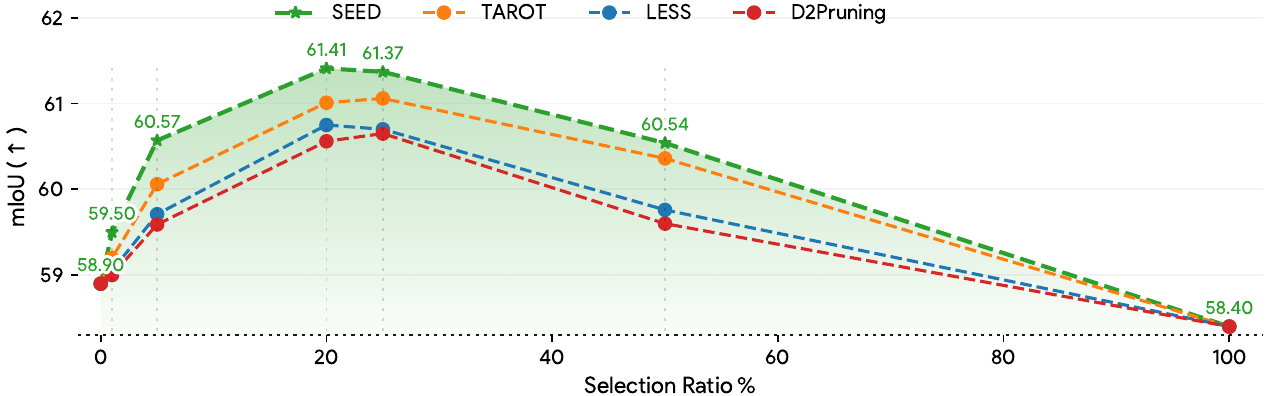}
    \vspace{-1mm}
    \caption{
        \textbf{Performance of SEED on scratch training for semantic segmentation.} Result curves of DeepLabV3 \cite{chen2017rethinking} trained using different data selection methods. Best viewed in color.
    }
    \label{fig:seg_res}
    \vspace{-3.5mm}
\end{figure}

\subsection{SEED for Visual Instruction Tuning}
\textbf{Settings.} We adopt Qwen3VL-4B \cite{bai2025qwen3} for visual instruction tuning. We refresh the annotations of Honeybee-1M \cite{bansal2025honeybee} using Doubao-1.6-VL \cite{guo2025seed1} to build a higher-quality candidate pool for all methods to select $5\%$. We further release the top $20\%$ selected by SEED as \texttt{Honeybee-Remake-SEED-200K}. We evaluate on $7$ representative benchmarks as the target dataset. Further details are in Appendix~\ref{sec:more_vlm}.

\begin{table}[t]
\centering
\vspace{-5mm}
\caption{\textbf{Visual instruction tuning performance} with Qwen3VL-4B under $5\%$ selection ratio.
}
\vspace{0mm}
\footnotesize
\label{tab:vlm}
\renewcommand{\arraystretch}{1.2}
\setlength{\tabcolsep}{0.3mm}

\begin{tabular}{p{1.6cm}p{.75cm}|ccccccc|c}
    \toprule[0.95pt]
    \multicolumn{2}{c|}{\textbf{Method}}
    & \scshape\textbf{Mathvista} 
    & \scshape\textbf{RealWorld} 
    & \scshape\textbf{MathVerse} 
    & \scshape\textbf{Hallusion} 
    & \begin{tabular}{c}
    \scshape\textbf{CharXiv}\\[-0.8mm]
    {\scriptsize (RQ/DQ)}
    \end{tabular}
    & \scshape\textbf{SimpleVQA} 
    & \scshape\textbf{MME}
    & \scshape\textbf{Rel. Avg.} \\
    
    \cmidrule(lr){1-2}
    \cmidrule(lr){3-9}
    \cmidrule(lr){10-10}
    
    \scshape\textbf{Base} & \texttt{\scriptsize{(0\%)}} & 72.4 & 71.2 & 41.2 & 56.7 & 41.2 / 80.7 & 47.9 & 2279.5 & 61.6 \\
    
    \rowcolor{mygray}
    \scshape\textbf{Full} & \texttt{\scriptsize{(100\%)}} & 78.4 & 73.7 & 51.8 & 55.9 & 50.8 / 75.8 & 47.5 & 2429.7 & 65.1  \\
    
    \midrule[0.6pt]
    
    \scshape\textbf{Random} & & 72.6 & 72.4 & 46.7 & 54.5 & 43.0 / 74.9 & 47.9 & 2390.2 & 62.2  \\
    
    \scshape\textbf{LESS} \cite{xia2024less} 
    & \texttt{\scriptsize{ICML24}} & 74.6 & 71.8 & \textbf{51.6} & 56.2 & 46.7 / 76.7 & 47.4 & 2319.5 & 63.5  \\
    
    \scshape\textbf{Tailor} \cite{yu2025mastering} 
    & \texttt{\scriptsize{ICCV25}} & 74.3 & 72.8 & 49.8 & 55.3 & \textbf{52.8} / 75.0 & 47.3 & 2412.0 & 64.2  \\
    
    \rowcolor{mygray}
    \scshape\textbf{SEED} & & \textbf{76.2} & \textbf{73.7} & 51.4 & \textbf{57.3} & 47.8 / \textbf{79.3} & \textbf{48.3} & \textbf{2477.4} & \textbf{65.4} \\
    
    \bottomrule[0.95pt]
\end{tabular}
\vspace{-0.5mm}
\end{table}

\textbf{Main Results.} 
As shown in Table~\ref{tab:vlm}, SEED demonstrates remarkable efficiency by achieving a state-of-the-art average score of $65.4$ with merely $5\%$ of the training data, outperforming the full-data baseline by $0.3$ points and DataTailor~\cite{yu2025mastering} by \textbf{1.2} points across multiple advanced benchmarks.

\begin{figure}[t]
    \centering
    \vspace{-1.5mm}
    \includegraphics[width=1\linewidth]{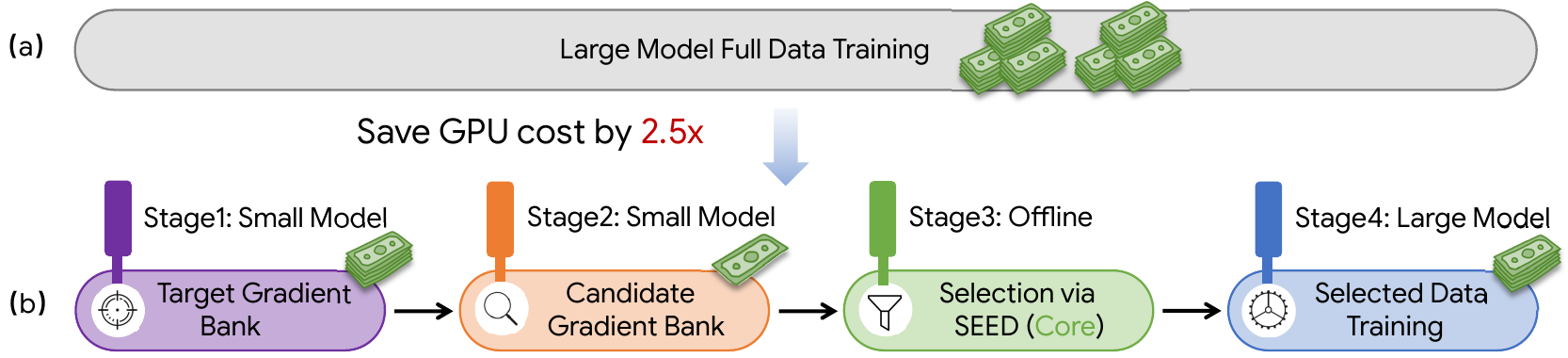}
    \vspace{-3.5mm}
    \caption{
        \textbf{Efficient fine-tuning via SEED-guided proxy model selection.} Statistics are reported on Phi-4-14B, where SEED-T achieves comparable accuracy while reducing GPU cost by 2.5$\times$.
    }
    \label{fig:proxy_pipeline}
    \vspace{-4.5mm}
\end{figure}

\begin{table}[t]
    \centering
	\caption{\textbf{Performance of SEED under proxy-driven selection settings.} Each group pairs a smaller proxy model with a larger target model.
    % , with progressively increasing scale gaps across heterogeneous model settings and varying data selection ratios. 
    SEED-T denotes the use of the transfer strategy.}
    \begin{minipage}[t]{0.48\textwidth}
	\renewcommand\arraystretch{1.2}
	\setlength\tabcolsep{1.4mm}
	\centering
	\vspace{0mm}
	\label{tab:det}
	\footnotesize
	%\scriptsize
	%\normalsize
    \begin{tabular}{l|ccc|c}
        \toprule[0.95pt]
        \textbf{Method} & \scshape\textbf{TyDiQA} & \scshape\textbf{MMLU} & \scshape\textbf{BBH} & \scshape\textbf{Avg.} \\
        
        \midrule[0.6pt]
        \multicolumn{5}{l}{\scriptsize{Proxy Model: \raisebox{-0.25\height}{\includegraphics[height=1.3em]{fig/qwen.png}}\hspace{0.2em}Qwen3-\textbf{4B}} (Scale: \textbf{2.00}$\times$)} \\
        \scriptsize{\raisebox{-0.25\height}{\includegraphics[height=1.3em]{fig/qwen.png}}\hspace{0.2em}Qwen3-\textbf{8B}} & 36.5 & 72.7 & 78.3 & 62.5\\
        \scshape\textbf{Full} 100\% & 52.9 & 75.6 & 73.1 & 67.2\\
        \scshape\textbf{Random} 1\% & 54.0 & 72.9 & 78.6 & \underline{68.5}\\
        \rowcolor{mygray}
        \scshape\textbf{SEED}-T 1\% & 55.9 & 75.2 & 79.5 & \textbf{70.2}\\
        
        \midrule[0.6pt]
        \multicolumn{5}{l}{\scriptsize{Proxy Model: \raisebox{-0.25\height}{\includegraphics[height=1.05em]{fig/llama.png}}\hspace{0.2em}LLaMA2-\textbf{7B}} (Scale: \textbf{5.14}$\times$)} \\
        \scriptsize{\raisebox{-0.25\height}{\includegraphics[height=1.05em]{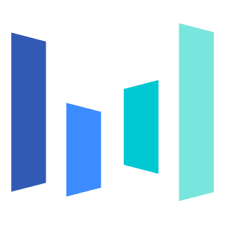}}\hspace{0.2em}Seed-OSS-\textbf{36B}} & 54.5 & 81.0 & 71.5 & 69.0 \\
        \scshape\textbf{Full} 100\% & 59.0 & 80.5 & 72.1 & 70.5 \\
        \scshape\textbf{Random} 5\% & 59.5 & 81.1 & 71.5 & \underline{70.7}\\
        \rowcolor{mygray}
        \scshape\textbf{SEED}-T 5\% & 60.5 & 83.3 & 72.4 & \textbf{72.1}\\
        
        \bottomrule[0.95pt]
    \end{tabular}
    \end{minipage}
    \hspace{2mm}
    \begin{minipage}[t]{0.48\textwidth}
    \renewcommand\arraystretch{1.2}
	\setlength\tabcolsep{1.4mm}
	\centering
	%\begin{center}
	\vspace{0mm}
	\label{tab:det2}
	\footnotesize
	%\scriptsize
	%\normalsize
    \begin{tabular}{l|ccc|c}
        \toprule[0.95pt]
        \textbf{Method} & \scshape\textbf{TyDiQA} & \scshape\textbf{MMLU} & \scshape\textbf{BBH} & \scshape\textbf{Avg.} \\
     
        \midrule[0.6pt]
        \multicolumn{5}{l}{\scriptsize{Proxy Model: \raisebox{-0.25\height}{\includegraphics[height=1.3em]{fig/qwen.png}}\hspace{0.2em}Qwen3-\textbf{1.7B}} (Scale: \textbf{8.24}$\times$)} \\
        \scriptsize{\raisebox{-0.25\height}{\includegraphics[height=1.05em]{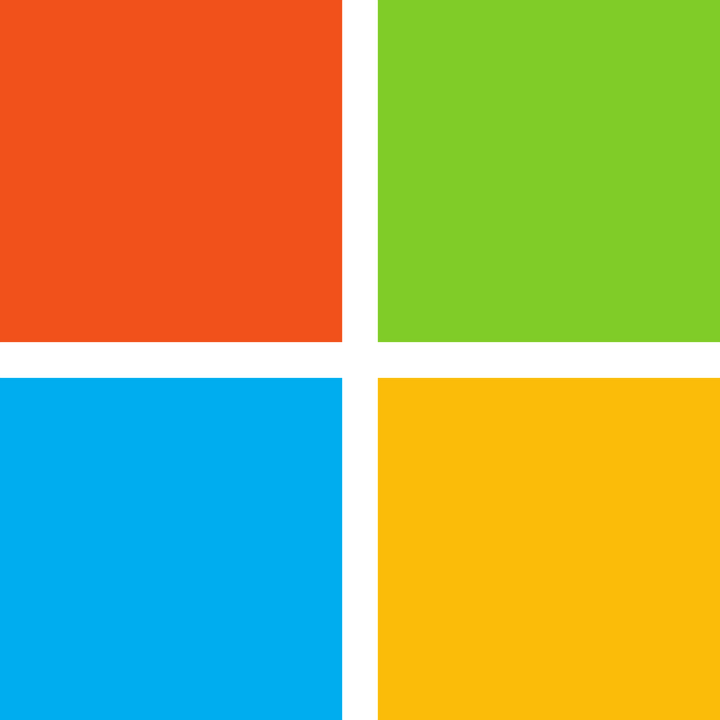}}\hspace{0.2em}Phi-4-\textbf{14B}} & 49.4 & 80.3 & 69.0 & 66.2\\
        \scshape\textbf{Full} 100\% & 54.1 & 82.0 & 73.1 & \underline{69.7}\\
        \scshape\textbf{Random} 5\% & 53.3 & 89.7 & 71.4 & 68.5\\
        \rowcolor{mygray}
        \scshape\textbf{SEED}-T 5\% & 55.8 & 83.2 & 73.2 & \textbf{70.7}\\

        \midrule[0.6pt]
        \multicolumn{5}{l}{\scriptsize{Proxy Model: \raisebox{-0.25\height}{\includegraphics[height=1.05em]{fig/llama.png}}\hspace{0.2em}LLaMA2-\textbf{7B}} (Scale: \textbf{10.00}$\times$)} \\
        \scriptsize{\raisebox{-0.25\height}{\includegraphics[height=1.05em]{fig/llama.png}}\hspace{0.2em}LLaMA2-\textbf{70B}} & 41.3 & 68.2 & 65.4 & 58.3\\
        \scshape\textbf{Full} 100\% & 53.0 & 69.4 & 69.2 & \textbf{63.9} \\
        \scshape\textbf{Random} 1K & 46.8 & 68.3 & 66.0 & 60.4 \\
        \rowcolor{mygray}
        \scshape\textbf{SEED}-T 1K & 51.2 & 69.7 & 67.1 & \underline{62.7} \\
        
        \bottomrule[0.95pt]
    \end{tabular}
    \end{minipage}
    \vspace{-2mm}
\end{table}

\subsection{SEED for Semantic Segmentation}

\textbf{Settings.}
We adopt the TAROT~\cite{feng2025tarot} setting, using GTA5~\cite{richter2016playing} as the source and Cityscapes~\cite{cordts2016cityscapes} as the target for training and evaluation. We utilize DeepLabV3~\cite{chen2017rethinking} with a ResNet-50~\cite{he2016deep} backbone, extracting gradients from the last four checkpoints to guide $1\%$, $5\%$, $20\%$, $25\%$, and $50\%$ selection. We use mean Intersection over Union (mIoU) as the primary evaluation metric.

% \textbf{Evaluation Metric.}
% We use mean Intersection over Union (mIoU) as the primary evaluation metric.

\textbf{Main Results.} Results are summarized in Figure~\ref{fig:seg_res}.
Among the evaluated methods, SEED consistently maintains the highest mIoU across all selection ratios, achieving its peak of $61.41$ mIoU at the $20\%$ mark, a substantial gain of $3.01$ over the full-data baseline.
% Notably, unlike the data-scaling trends typical in LLM research, real-world urban data contains substantial noise, where full-data training can hurt performance, making targeted selection critical over brute-force data accumulation.

\subsection{SEED with Small Proxies}
For single-epoch training of large-scale LLMs, computing influence scores over the entire candidate set often exceeds the training cost itself. To ensure efficiency, we propose a proxy-driven selection strategy (Figure~\ref{fig:proxy_pipeline}), utilizing a small proxy model to identify high-value data for a larger target model.

\textbf{Settings.}
For proxy models, we employ LLaMA2-7B \cite{touvron2023llama}, Qwen3-1.7B, and Qwen3-4B \cite{yang2025qwen3}. For target models, we consider LLaMA3-8B \cite{dubey2024llama}, LLaMA2-13B, and LLaMA2-70B \cite{touvron2023llama}, Qwen3-8B \cite{yang2025qwen3}, DeepSeek-LLM-7B \cite{bi2024deepseek}, Phi-4-14B \cite{abdin2024phi}, and Seed-OSS-36B \cite{seed2025seed-oss}. 
% This diverse model pool allows us to assess cross-family transferability and scalability under varying capacity gaps. 
We report four representative cases here and provide all eight results in the Appendix~\ref{sec:transfer_full}.

\textbf{Main Results.}
% In Table~\ref{tab:det}, SEED-T frequently \textbf{surpasses full-data fine-tuning} using only a small fraction of the training data and a smaller proxy model, with gains of up to \textbf{3.0} points on Qwen3-8B and \textbf{1.6} points on Seed-OSS-36B. Even at the extreme 10.00$\times$ scale gap with only 1K samples, SEED-T approaches the full-data upper bound within 1.2 points while outperforming random selection by 2.3 points. These results demonstrate that large language models can be efficiently trained on carefully selected subsets via SEED, without requiring a same-scale proxy model.
In Table~\ref{tab:det}, SEED frequently \textbf{surpasses full-data fine-tuning} via minimal samples and smaller proxies, with gains of up to \textbf{3.0} points on Qwen3-8B and \textbf{1.6} points on Seed-OSS-36B. Remarkably, with a 10$\times$ scale gap and only 1K samples, SEED outperforms random selection by $2.3$, approaching full-data performance within $1.2$. The results validate that large models can be efficiently trained on carefully selected subsets via SEED, without requiring a same-scale proxy model.

\section{Analysis}

% \setlength{\columnsep}{5.5mm}
% \begin{wrapfigure}{r}{0.4\textwidth}
%     \vspace{-4mm}
%     \begin{center}
%         \includegraphics[width=\linewidth]{fig/trade-off.pdf}
%     \end{center}
%     \vspace{-4mm}
%     \caption{\textbf{Curves of performance vs.\ training cost on Phi-4-14B.} SEED-T achieves better to full-data training.}
%     \label{fig:cost}
%     \vspace{-2mm}
% \end{wrapfigure}
\begin{figure}[t]
    \centering
    \setlength{\tabcolsep}{3pt}

    \begin{minipage}{0.53\linewidth}
    \centering
    % \scriptsize
    \renewcommand{\arraystretch}{1.7}
    \captionof{table}{\textbf{Ablation of components in SEED,} using the proposed node and edge formulations. Results are reported on LLaMA3-8B with a $1\%$ data ratio.}
    \label{tab:ab}
    \resizebox{\linewidth}{!}{
        \definecolor{mygraytext}{gray}{.75}
\newcommand{\graytext}[1]{\textcolor{gray}{\raisebox{0.15ex}{#1}}}

% \begin{table}[t]
%   \caption{\textbf{Ablation of components in SEED.}}
%   \vspace{-1mm}
%   \renewcommand{\arraystretch}{1.25}
%   \setlength{\tabcolsep}{0.7mm}
%   \label{tab:ab}
%   \centering
\begin{tabular}{ccc| ccc}
    \toprule
    + (\textit{\textbf{MWIS}}) & + (\textit{\textbf{Nodes}}) & + (\textit{\textbf{Edges}}) & \scshape\textbf{TyDiQA} & \scshape\textbf{MMLU} & \scshape\textbf{BBH}\\
    \midrule
    \graytext{\xmark} & \graytext{\xmark} & \graytext{\xmark} & 54.1 & 65.6 & 67.8  \\
    
    \cmark & \graytext{\xmark} & \graytext{\xmark} & 54.7 & 65.2 & 68.0 \\
    
    \graytext{\cmark} & \cmark & \graytext{\xmark} & 55.7 & 65.9 & 69.1 \\
    
    \graytext{\cmark} & \graytext{\xmark} & \cmark & 55.5 & 66.0 & 68.7 \\
    \rowcolor{mygray}
    \graytext{\cmark} & \graytext{\cmark} & \graytext{\cmark} & 56.1 & 66.6 & 69.9 \\
    
    \bottomrule
\end{tabular}

    }
        \vspace{0mm}
    \end{minipage}
    \hfill
    \begin{minipage}{0.43\linewidth}
        \centering
        \includegraphics[width=\linewidth]{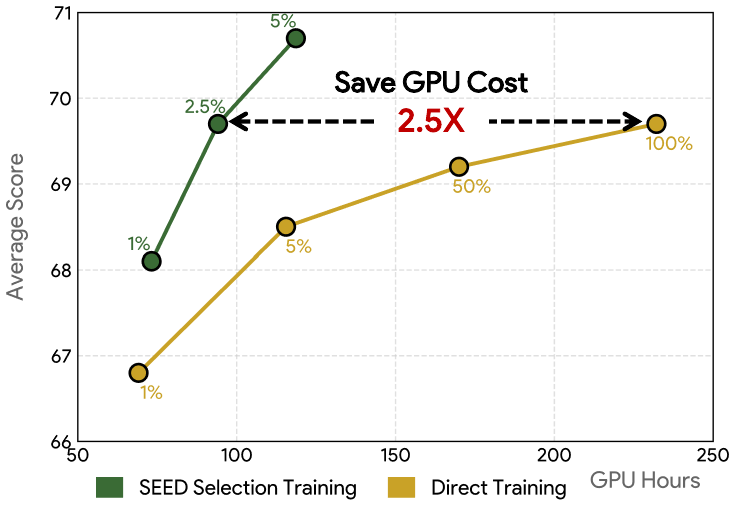}
        \vspace{-5mm}
        \captionof{figure}{\textbf{Curve of performance vs.\ GPU training cost on Phi-4-14B.}}
        \label{fig:cost}
    \end{minipage}
    \vspace{-6mm}
\end{figure}

\subsection{Ablation Study on SEED Components}
We conduct an ablation study to quantify the contribution of each component in SEED (Table~\ref{tab:ab}).
\textbf{Influence Score.}
Provides a strong baseline over random sampling, establishing a relevance signal.
\textbf{+ MWIS.}
Introducing structural diversity yields gains, but a slight drop on MMLU suggests limited robustness across heterogeneous scenarios.
\textbf{+ Node Representation.}
Improved representations bring consistent gains ($+0.9$ avg), highlighting the importance of subspace-calibrated relevance estimation.
\textbf{+ Edge Construction.}
Refined edges further improve performance ($+0.6$ avg) by correcting local scaling and aligning neighborhood distributions (Figure~\ref{fig:local_scale_motivation}(c)).
Overall, SEED improves by $+1.7$ over the influence baseline ($64.2$ vs.\ $62.5$), demonstrating the complementary effects of all components.

\subsection{Practical Efficiency Analysis}
While the results in §\ref{sec:exp} demonstrate the accuracy advantages of SEED, we further analyze its practical efficiency in real-world applications. Figure~\ref{fig:cost} illustrates the performance–cost trade-off on Phi-4-14B \cite{abdin2024phi}.
% At matching accuracy levels, SEED requires only $2.5\%$ of the training data, reducing GPU costs by $\mathbf{2.5\times}$. Furthermore, SEED at $5\%$ selection achieves a $70.7$ score, surpassing the full-data baseline at less than half the computational budget. 
% In practice, this translates to significant reductions in both training time and computational cost, which matter when fine-tuning large models at scale. These findings suggest that careful data selection via SEED can substitute for brute-force data accumulation, even when only a small proxy model is available.
Notably, SEED achieves comparable accuracy to full-data training using only $2.5\%$ of the samples, yielding a $\mathbf{2.5\times}$ reduction in GPU cost, corresponding to savings of approximately $138$ A100 GPU hours (\textasciitilde\$$\textbf{500}$ at Google Cloud on-demand pricing, \$$3.67$ per hour). Furthermore, at a $5\%$ selection ratio, SEED reaches a $70.7$ average score, outperforming the full-data baseline while halving the computational budget. These results suggest that SEED-driven data selection is a superior alternative to brute-force training, especially in resource-constrained fine-tuning scenarios.

\subsection{Sensitivity Analysis}
As introduced in \S\ref{sec:Edge_Construction}, SEED employs the $k$-th nearest neighbor to estimate local density $\sigma_i$ for adaptive edge construction, where a larger $k$ captures a broader neighborhood context but may smooth out fine-grained density differences. We investigate how the choice of $k$ and scaling factor $\alpha$ affects data selection performance on instruction tuning tasks. Results in 
Appendix~\ref{sec:Sensitivity_Analysis} show that SEED remains robust across a wide 
range of $k$ and $\alpha$ values, validating the stability of our design choices.

\subsection{Visualization}

\setlength{\columnsep}{5.5mm}
\begin{wrapfigure}{r}{0.42\textwidth}
    \vspace{-11mm}
    \begin{center}
        \includegraphics[width=\linewidth]{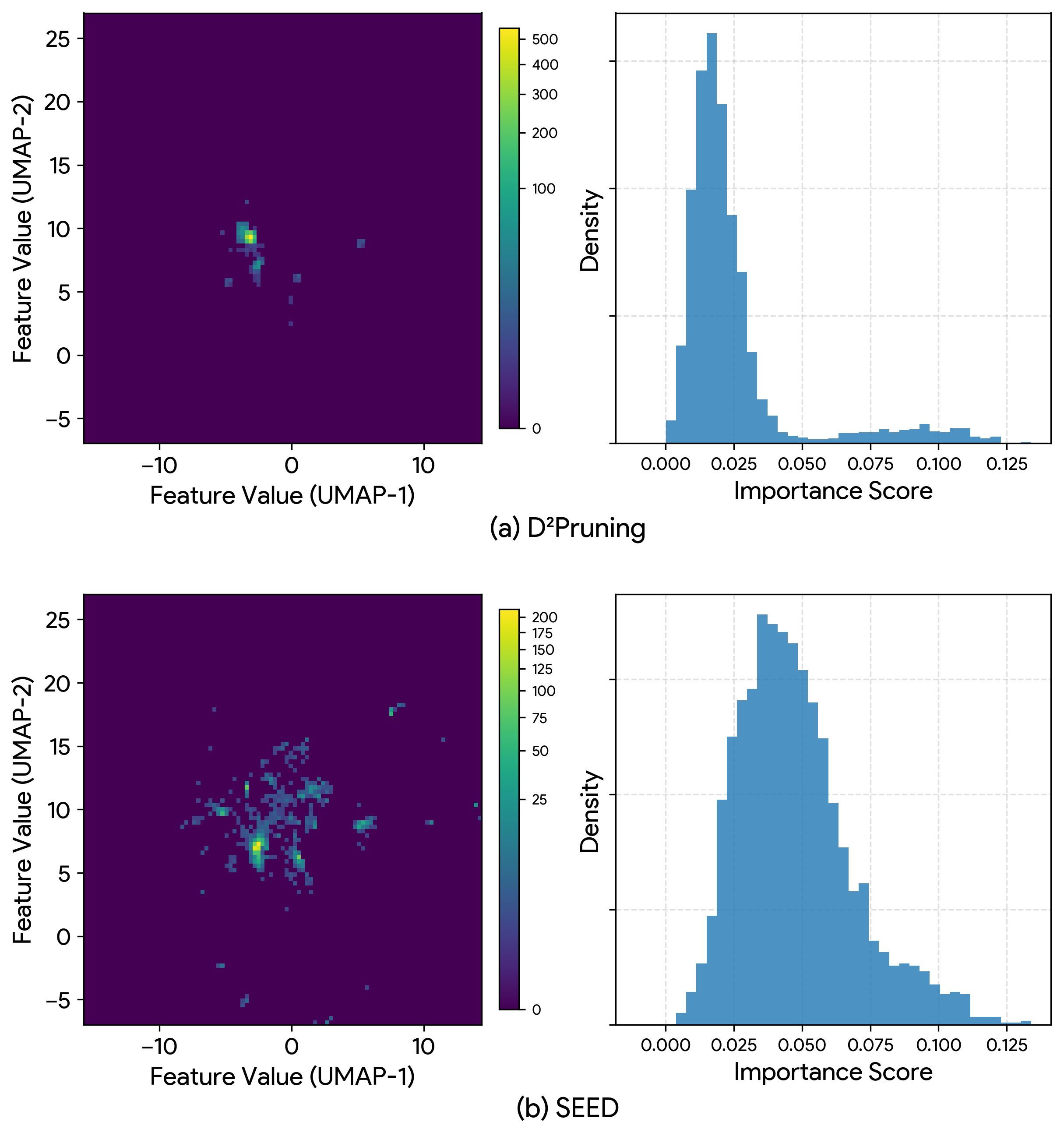}
    \end{center}
    \vspace{-4mm}
    \caption{\textbf{Visualization} of data manifolds (left) and influence scores (right).}
    \label{fig:vis}
    \vspace{-11.5mm}
\end{wrapfigure}

To investigate the selection behavior, we visualize projected manifolds via UMAP~\cite{mcinnes2018umap} and influence score distributions in Figure~\ref{fig:vis} on the instruction tuning task. 
Results reveal that D$^2$Pruning's selections collapse into a highly concentrated region due to global threshold bias, yielding spatially redundant subsets with left-skewed score distributions. In contrast, SEED achieves broader and more uniform coverage across the manifold while selecting samples with a more balanced score distribution and fewer low-quality outliers.
\section{Conclusion}
We present SEED, a robust data selection pipeline that formulates subset selection as a Weighted Independent Set problem on a similarity graph. To address two fundamental failure modes of naive WIS, we introduce node value calibration via mutual influence subspace restriction and local scale normalization for density-adaptive edge construction. 
% Together, these refinements yield compact subsets that are simultaneously high-quality and diverse.
We further release \texttt{Honeybee-Remake-SEED-200K}, a high-quality multimodal dataset curated by SEED.
Extensive experiments on instruction tuning, visual instruction tuning, and semantic segmentation demonstrate that SEED consistently outperforms state-of-the-art methods across diverse 
model families and selection ratios, while reducing GPU costs by up to 2.5$\times$ via proxy-driven selection.

\newpage
{
\small
\bibliography{main}
\bibliographystyle{abbrv}
}

%%%%%%%%%%%%%%%%%%%%%%%%%%%%%%%%%%%%%%%%%%%%%%%%%%%%%%%%%%%%

\addtocontents{toc}{\protect\setcounter{tocdepth}{2}}
\clearpage
\setcounter{page}{1}
\appendix

\setcounter{section}{0}
\renewcommand{\thesection}{\Alph{section}}
\tableofcontents

\section{Related Work}

\subsection{Data Selection or Pruning}
Coreset selection aims to identify a compact subset for optimal downstream model performance for efficient training \cite{wei2015submodular, sener2018active, maharana2023d2, xie2023dsir, liu2024deita, xia2024less, tandata, feng2025tarot, matask, zhang2025d3, dai2025improving}. Existing approaches broadly can fall into three categories: \textit{quality}-driven, \textit{diversity}-driven, and \textit{hybrid} methods.

\textbf{Quality-based Methods.}
Quality-driven approaches prioritize samples by heuristic difficulty~\cite{maharana2023d2}, training loss~\cite{paul2021deep}, or estimated impact on model behavior. A prominent line of work uses influence functions~\cite{koh2017understanding} to approximate a sample's effect on validation loss. Exact Hessian computations being costly, later methods adopt efficient trajectory-based proxies like TracIn~\cite{pruthi2020estimating} and LESS~\cite{xia2024less}. While effective at identifying high-impact samples, pure importance scoring often selects redundant samples clustered around high-density regions, thereby limiting coverage of the data distribution.

\textbf{Diversity-driven Methods.}
To ensure broad coverage, diversity-driven techniques maximize feature-space dispersion, with classical strategies including $k$-center clustering~\cite{sener2018active} and facility location~\cite{wei2015submodular}. For modern foundation models, prototype-based selection~\cite{liu2024deita} and semantic deduplication~\cite{abbas2023semdedup} show that removing near-duplicates cuts training costs without significantly harming accuracy. However, a pure diversity focus risks retaining noisy outliers, particularly in web-scale.

\textbf{Hybrid Methods.}
Recent work emphasizes balancing quality and diversity~\cite{maharana2023d2, tandata, dai2025improving, zhang2025d3}. A natural formulation constructs a similarity graph where nodes are samples and edges encode redundancy, reducing selection to graph optimization. For example, $\text{D}^2$Pruning~\cite{maharana2023d2} and DSIR~\cite{xie2023dsir} implicitly combine both criteria. However, prior graph-based approaches typically rely on complex graph structures with weighted edges, and overlook two critical failures: (1) \textit{representation miscalibration}: noise in gradients distorts node importance; and (2) \textit{structural imbalance}: heterogeneous domain densities cause global thresholds to suppress samples in dense regions. Our method addresses these issues via mutual influence subspace and local scale normalization under a weighted independent set.

\subsection{Weighted Independent Set for Data Selection}

The weighted independent set formulation provides a principled graph-theoretic framework for data selection, where nodes represent samples weighted by importance and edges encode mutual influence. The most related work is GIST~\cite{fahrbach2025gist}, which solves a diversity-centric max-min diversification problem via independent sets on threshold graphs. However, GIST and SEED differ fundamentally in both \textit{problem definition} and \textit{applicable scope}, making direct experimental comparison inappropriate.

\textbf{Incompatible Objective Functions.} GIST optimizes $f(\mathcal{S}) = g(\mathcal{S}) + 
\mathrm{div}(\mathcal{S})$, where $\mathrm{div}(\mathcal{S}) = \min_{u,v \in \mathcal{S}, u 
\neq v} \mathrm{dist}(u, v)$ is a purely geometric diversity term defined over a metric space. 
Node weights in GIST are geometry-agnostic: all points are treated as equally weighted in a 
uniform metric space, with no mechanism to incorporate task-specific quality signals. In 
contrast, SEED assigns each node a trajectory-based influence score derived from a mutual 
gradient subspace (§\ref{sec:node_calibration}), grounding node importance in task-relevant 
learning dynamics rather than surface-level geometry.

\textbf{Divergent Structural Assumptions.} GIST assumes a domain-homogeneous embedding space 
and applies a global distance threshold across all nodes. As demonstrated in §\ref{sec:Edge_Construction} and Figure~\ref{fig:local_scale_motivation}, this is a critical failure mode under heterogeneous domain distributions: dense regions accumulate excessive edges and become over-suppressed, while sparse regions remain under-connected and tend to be over-selected. SEED explicitly addresses this via local scale normalization, which adapts edge thresholds to local neighborhood density.

\textbf{Incompatible Experimental Protocols.} GIST is validated on a single-shot subset selection for image classification~(ImageNet), with no established protocol for instruction tuning or visual instruction tuning. Adapting GIST to our setting would require non-trivial design choices in defining the utility function $g(\mathcal{S})$, introducing confounds that would render any comparison unfair to both methods.

Correlation-constrained variable selection~\cite{colling2025corrselect} employs a similar 
independent set formulation, but likewise optimizes geometry-based diversity without 
task-specific quality signals. Together, these works confirm that SEED is the first weighted 
independent set framework for \textit{targeted}, large-scale data selection across LLMs, VLMs, 
and segmentation tasks---one that jointly addresses representation miscalibration and 
structural graph imbalance under realistic cross-domain conditions.

\section{Implementation Details}
\label{sec:implement}   

\subsection{Efficient Implementation of WIS}
\label{sec:app_mwis_impl}                            

To solve the \textit{Weighted Independent Set} (WIS) problem efficiently on large-scale datasets, we implement a two-step greedy approximation algorithm optimized for GPUs.

\textbf{(1) Sparse Conflict Graph Construction.} 
We utilize FAISS \cite{johnson2019billion} on GPUs to compute pairwise influence between normalized node embeddings. To handle large corpus sizes gracefully, the search is executed in batches to retrieve the top-$k$ nearest neighbors for each node. An edge between two nodes is established in a sparse adjacency matrix if their similarity exceeds a dynamic threshold. Specifically, to adapt to varying local embedding densities, we employ a conservative local scaling strategy \cite{zelnik2004self}. For a given node $i$, the threshold is defined as $\max(\tau, \alpha \cdot \sigma_i)$, where $\tau$ is a global base threshold, $\sigma_i$ is the similarity of node $i$'s $k$-th nearest neighbor, and $\alpha \in [0, 1]$ is a scaling factor. This adaptive formulation prevents overly dense subgraphs in regions with high semantic overlap.

\noindent\textbf{(2) Greedy Selection via Max-Heap.}
Given the constructed sparse conflict graph and the node influence scores (weights), we initialize a \textit{max-heap} to prioritize nodes based on their exact weights. In each iteration, we pop the node with the highest weight. If the node has not been marked as removed, it is added to the selected independent set $S$. Crucially, all its adjacent neighbors in the sparse graph are simultaneously marked as removed. This process continues until the heap is empty or the desired target data size is reached, ensuring a highly efficient selection phase.

\subsection{Training Configurations}
Table~\ref{tab:model_config} summarizes the training configurations of SEED across all tasks. For instruction tuning, we follow the experimental setup of LESS~\cite{xia2024less}, selecting from a candidate pool of 270K samples drawn from FLAN-V2, CoT, Dolly, and OpenAssistant-1. All LLM experiments adopt LoRA fine-tuning with rank 128 and alpha 512, projecting gradients to an 8192-dimensional subspace. For visual instruction tuning, we fine-tune Qwen3-VL 4B on Honeybee-Remake-SEED with full model training on A100 GPUs. For semantic segmentation, we train DeepLabV3 with a ResNet-50 backbone on RTX 4090 GPUs using polynomial learning rate decay.

\begin{table}[h]
    \centering
    \caption{
    \textbf{Training configurations of SEED} across LLM, VLM, and segmentation tasks.
    }
    \vspace{1mm}
    \setlength{\tabcolsep}{10pt}
    \renewcommand{\arraystretch}{1.3}
    \resizebox{\textwidth}{!}{%
    \begin{tabular}{@{}ll|ccc|c|c@{}}
        \toprule
        & \textbf{Tasks} & \multicolumn{3}{c|}{\textbf{LLM}} 
            & \textbf{VLM} 
            & \textbf{Segmentation} \\
        \cmidrule(l){1-2} \cmidrule(l){3-5} \cmidrule(l){6-6} \cmidrule(l){7-7}
        & \textbf{Models} & \textbf{LLaMA-2-7B} 
            & \textbf{LLaMA-3-8B} 
            & \textbf{Qwen3-4B} 
            & \textbf{Qwen3-VL} 
            & \textbf{DeepLabV3} \\
        \midrule 
        
        \multirow{2}{*}{\rotatebox[origin=c]{90}{\small \textit{Data}}}
        & \textbf{Dataset} 
        & Text & Text 
        & Text & Image, Text 
        & Image \\
        
        & \#Full Candidates 
        & 270679 & 270679 
        & 270679 & 1023915
        & 24966 \\

        & \#Target Samples & \multicolumn{3}{c|}{TyDiQA: 9, MMLU: 285, BBH: 81} & 5\% Full Targets & 2975\\
        
        \midrule
        
        \multirow{2}{*}{\rotatebox[origin=c]{90}{\small \textit{Model}}}
        & \textbf{Trainable} 
        & LoRA 
        & LoRA 
        & LoRA 
        & Full Model 
        & Full Model \\

        & \#Rank, Alpha
        & 128, 512 & 128, 512 
        & 128, 512 & -
        & - \\

        & \#Projector Dimension 
        & 8192 & 8192 
        & 8192 & - 
        & - \\

        & \#Tunable Parameters 
        & 134.2M & 109.1M
        & 94.4M &  4.4B
        & 42.0M \\
        
        \midrule 
        
        \multirow{3}{*}{\rotatebox[origin=c]{90}{\small \textit{Training}}}
        & \textbf{Batch Size} 
        & 1 & 1 & 1 & 8 
        & 16 \\

        & \textbf{Optimizer} & AdamW & AdamW & AdamW & AdamW & SGD \\
        
        & \textbf{LR Schedule} 
        & Linear & Linear & Linear
        & Cosine
        & Poly decay \\
        
        & \#LR
        & $2 \times 10^{-5}$
        & $2 \times 10^{-5}$
        & $2 \times 10^{-5}$
        & $1 \times 10^{-5}$
        & $1 \times 10^{-2}$ \\

        & \textbf{Epoch} & 4 & 4 & 4 & 2 & 40K iterations\\

        & \textbf{Hardware} 
        & A100 & A100 & A100 & A100 & RTX 4090 \\
        
        \bottomrule
    \end{tabular}
    }
    \label{tab:model_config}
\end{table}

\subsection{Evaluation Setup for Visual Instruction Tuning}
\label{sec:vlm_eval}
\paragraph{Benchmarks.}
We evaluate visual instruction tuning with seven representative multimodal benchmarks covering mathematical reasoning, real-world spatial understanding, diagram-based reasoning, hallucination diagnosis, chart understanding, factuality, and general perception. A detailed overview of these benchmarks is provided in the following Table~\ref{tab:vlm_benchmarks}. 

\begin{table}[h]
\centering
\caption{\textbf{Overview of visual instruction tuning evaluation benchmarks.}}
\vspace{1mm}
\label{tab:vlm_benchmarks}
\footnotesize
\renewcommand{\arraystretch}{1.3}
\setlength{\tabcolsep}{4mm}
\begin{tabular}{p{2.8cm}p{10.4cm}}
    \toprule
    \textbf{Benchmark} & \textbf{Description} \\
    \midrule
    MathVista~\cite{lu2024mathvista} 
        & Evaluates mathematical reasoning in visual contexts 
          (\texttt{MathVista\_mini} split). \\[2pt]
    RealWorldQA~\cite{xai2024grok15v} 
        & Assesses real-world spatial understanding via 
          image-grounded, verifiable QA pairs. \\[2pt]
    MathVerse~\cite{zhang2024mathverse} 
        & Probes whether MLLMs genuinely leverage visual 
          diagrams in multimodal math problem solving. \\[2pt]
    HallusionBench~\cite{guan2024hallusionbench} 
        & Diagnoses language hallucination and visual illusion 
          in image-context reasoning. \\[2pt]
    CharXiv (RQ/DQ)~\cite{wang2024charxiv} 
        & Benchmarks chart understanding on arXiv figures, 
          covering reasoning (RQ) and descriptive (DQ) questions. \\[2pt]
    SimpleVQA~\cite{cheng2025simplevqa} 
        & Measures multimodal factuality via short-answer questions 
          with LLM-as-a-judge scoring. \\[2pt]
    MME~\cite{fu2023mme} 
        & Comprehensively evaluates multimodal perception and 
          cognition across diverse subtasks. \\
    \bottomrule
\end{tabular}
\vspace{-1mm}
\end{table}

\paragraph{Evaluation Framework.}
We adopt the official VLMEvalKit~\cite{duan2024vlmevalkit} as the unified evaluation framework for inference, post-processing, and metric computation. For Qwen3-VL, we set \texttt{temperature=0.7}, \texttt{max\_new\_tokens=8192}, \texttt{repetition\_penalty=1.0}, \texttt{presence\_penalty=1.5}, \texttt{top\_p=0.8}, and \texttt{top\_k=20}. For LLaVA-OneVision-1.5, we use deterministic decoding with \texttt{do\_sample=false} and \texttt{max\_new\_tokens=512}. For benchmarks requiring model-based judging, we utilize \texttt{gpt-4.1-mini-2025-04-14} as the LLM judge.

\section{Wall-Clock Time Analysis}
\label{appendix:wallclock}
We analyze the wall-clock time of SEED across two stages: gradient computation and data selection. As shown in Table~\ref{tab:efficiency2}, gradient computation is shared across all influence-based methods and \textbf{constitutes the dominant cost}. The data selection stage of SEED requires only $3.7$ minutes on a candidate pool of 270K samples, which is significantly faster than TSDS (10 hours) due to its KDE-based near-quadratic bottleneck, and slightly longer than LESS and TAROT owing to the additional MWIS optimization. Crucially, both gradient computation and data selection are \textit{one-time} costs that can be amortized across multiple model families, as demonstrated by our proxy-driven selection experiments.

\begin{table}[h]
\centering
\vspace{-1mm}
\caption{\textbf{Wall-clock time comparison} of data selection methods on instruction tuning task ($270K$ candidates, 
A100 GPU). Gradient computation is a one-time cost across influence-based methods.}
\footnotesize
\renewcommand{\arraystretch}{1.5}
\setlength{\tabcolsep}{13mm}
\label{tab:efficiency2}
\begin{tabular}{lcc}
\toprule
\textbf{Method} & \textbf{Gradient Computation} & 
\textbf{Data Selection} \\
\midrule
LESS~\cite{xia2024less}       & 28 Hours & 46 Seconds \\
TSDS~\cite{liu2024tsds}          & 28 Hours & 10 Hours   \\
TAROT~\cite{feng2025tarot}    & 28 Hours & 118 Seconds \\
\textbf{SEED (Ours)}          & 28 Hours & 3.7 Minutes \\
\bottomrule
\end{tabular}
\end{table}

\section{Multimodal Training Dataset: Honeybee-Remake-SEED-200K}
\label{appendix:dataset}

\paragraph{Base Corpus.}
We build upon Honeybee-1M~\cite{bansal2025honeybee}, a large-scale multimodal dataset originally designed for 
vision-language reasoning. We use the raw data directly without additional preprocessing.

\paragraph{Annotation Refresh.}
To improve annotation quality, we re-annotate the full Honeybee-1M corpus using Doubao-1.6-VL~\cite{guo2025seed1}, a state-of-the-art vision-language model, replacing the original ground-truth labels with higher-quality model-generated annotations.

\paragraph{Data Selection via SEED Voting.}
We apply SEED independently for each of the 7 target benchmarks, where each benchmark selects its own top-$K_b$ 
candidates based on the MWIS objective. Each time a candidate $z_i$ is selected by a benchmark, it receives 
one vote. The final score of each candidate is its accumulated vote count across all benchmarks:
\begin{equation}
    v_i = \sum_{b=1}^{7} 
    \mathbf{1}\left[z_i \in \mathcal{S}^*_b\right],
\end{equation}
where $\mathcal{S}^*_b$ denotes the subset selected by SEED for the $b$-th benchmark. Candidates are ranked by 
$v_i$ in descending order, and the top 20\% are retained, yielding the final \texttt{Honeybee-Remake-SEED-200K} 
dataset of approximately 200K multimodal instruction pairs.

\paragraph{Training Results.}
% 放20% ratio的实验结果即可
Table~\ref{tab:data_res} evaluates \texttt{Honeybee-Remake-SEED-200K} on two 4B-scale VLMs. Compared with random selection, our data improves the average 
score by $2.9$ points on Qwen3-VL and $4.0$ points on LLaVA-OneVision-1.5. It remains close to 
full-data training on Qwen3-VL and further surpasses the full-data baseline on LLaVA-OneVision-1.5 
($59.9$ vs.\ $59.3$), indicating that SEED voting produces a compact subset with good cross-backbone 
transferability.

\begin{table}[h]
\centering
\vspace{0mm}
\caption{\textbf{Performance comparison of different training data on multiple VLM benchmarks.} Base denotes training without any additional data (0\%), Full uses the complete dataset (100\%), Random applies random selection (20\%), and Ours refers to our proposed Honeybee-Remake-SEED-200K.}
\vspace{0mm}
\footnotesize
\label{tab:data_res}
\renewcommand{\arraystretch}{1.2}
\setlength{\tabcolsep}{0.7mm}

\begin{tabular}{p{1.1cm}p{.75cm}|ccccccc|c}
    \toprule[0.95pt]
    \multicolumn{2}{c|}{\textbf{Training Data}}
    & \scshape\textbf{Mathvista} 
    & \scshape\textbf{RealWorld} 
    & \scshape\textbf{MathVerse} 
    & \scshape\textbf{Hallusion} 
    & \begin{tabular}{c}
    \scshape\textbf{CharXiv}\\[-0.8mm]
    {\scriptsize (RQ/DQ)}
    \end{tabular}
    & \scshape\textbf{SimpleVQA} 
    & \scshape\textbf{MME}
    & \scshape\textbf{Avg.} \\
    
    \cmidrule(lr){1-2}
    \cmidrule(lr){3-9}
    \cmidrule(lr){10-10}
    
    \rowcolor{mygray}
    \multicolumn{10}{c}{\textit{Qwen3-VL} (4B)} \\
    \scshape\textbf{Base} & \texttt{\scriptsize{(0\%)}} & 72.4 & 71.2 & 41.2 & 56.7 & 41.2 / \textbf{80.7} & 47.9 & 2279.5 & 61.6 \\
    
    \scshape\textbf{Full} & \texttt{\scriptsize{(100\%)}} & \textbf{78.4} & \textbf{73.7} & \textbf{51.8} & 55.9 & 50.8 / 75.8 & 47.5 & \textbf{2429.7} & \textbf{65.1}  \\
  
    \scshape\textbf{Random} & \texttt{\scriptsize{(20\%)}} & 73.6 & 71.8 & 49.4 & 53.3 & 47.3 / 74.8 & 48.8 & 2401.0 & 63.1 \\

    \scshape\textbf{Ours} & \texttt{\scriptsize{(20\%)}} & 74.4 & 73.2 & 48.4 & \textbf{56.8} & \textbf{51.5} / 76.9 & \textbf{49.0} & 2398.1 & 64.5 \\
    
    \rowcolor{mygray}
    \multicolumn{10}{c}{\textit{LLaVA-OneVision-1.5} (4B)} \\
    \scshape\textbf{Base} & \texttt{\scriptsize{(0\%)}} 
    & 67.7 & \textbf{71.4} & 39.6 & 46.7 & 37.7 / 67.8 & 36.3 & 2248.7 & 55.9 \\
    
    \scshape\textbf{Full} & \texttt{\scriptsize{(100\%)}} 
    & 69.1 & 69.4 & 38.9 & \textbf{53.2} & \textbf{53.5} / 73.6 & 36.9 & 2244.7 & 59.3 \\
    
    \scshape\textbf{Random} & \texttt{\scriptsize{(20\%)}} 
    & 66.4 & 69.0 & 37.0 & 51.6 & 52.0 / 72.2 & \textbf{39.2} & 2256.7 & 58.5 \\
    
    \scshape\textbf{Ours} & \texttt{\scriptsize{(20\%)}} 
    & \textbf{70.5} & 68.8 & \textbf{44.6} & 52.4 & 48.5 / \textbf{75.0} & 38.6 & \textbf{2260.8} & \textbf{59.9} \\
    
    \bottomrule[0.95pt]
\end{tabular}
\vspace{-0.5mm}
\end{table}

\section{More Experiments}
\subsection{More Instruction Tuning Experiments}
To further assess the generalization capability of SEED, we conduct additional experiments on four out-of-domain benchmarks: SIQA~\cite{sap2019social} (0-shot), HumanEval~\cite{chen2021evaluating} (0-shot), GSM8K~\cite{cobbe2021training} (4-shot), and CLUE-C3~\cite{sun2020investigating} (0-shot). These datasets span diverse domains, including commonsense, code, math, and Chinese, none of which overlap with the instruction tuning candidate pool. We randomly sample $10$ examples from each dataset as the target set for influence estimation.

Results in Table~\ref{tab:ood} show that SEED generalizes effectively to out-of-domain benchmarks, outperforming full-data training on SIQA ($+2.4$), HumanEval ($+2.7$), and CLUE-C3 ($+3.0$) using only $5\%$ of the data, spanning domains entirely absent from the candidate pool. However, SEED underperforms full-data training on GSM8K, likely because mathematical reasoning requires dense coverage of problem-solving patterns that influence-based selection on a small target set cannot fully capture. Overall, these results suggest that subspace-calibrated influence estimation transfers well across diverse domains, while remaining sensitive to tasks that demand broad procedural coverage.

\begin{table}[h]
\centering
\vspace{0mm}
\caption{\textbf{Out-of-domain generalization of SEED on LLaMA2-7B under $5\%$ selection ratio.}}
\vspace{0mm}
\footnotesize
\label{tab:ood}
\renewcommand{\arraystretch}{1.2}
\setlength{\tabcolsep}{6mm}

\begin{tabular}{p{1.6cm}p{.75cm}|cccc}
    \toprule[0.95pt]
    \multicolumn{2}{c|}{\textbf{LLaMA2-7B}}
    & \scshape\textbf{SIQA} 
    & \scshape\textbf{HumanEval} 
    & \scshape\textbf{GSM8K} 
    & \scshape\textbf{CLUE-C3} \\
    
    \cmidrule(lr){1-2}
    \cmidrule(lr){3-6}
    
    \scshape\textbf{Base} & \texttt{\scriptsize{(0\%)}} & 48.3 & 14.6 & 16.5 & 42.1 \\
    
    \scshape\textbf{Full} & \texttt{\scriptsize{(100\%)}} & 59.4 & 13.8 & \textbf{19.7} & 48.1\\
  
    \scshape\textbf{Random} & \texttt{\scriptsize{(5\%)}} & 55.6 & 11.0 & 17.9 & 45.3\\

    \scshape\textbf{SEED} & \texttt{\scriptsize{(5\%)}} & \textbf{61.8} & \textbf{16.5} & 19.3 & \textbf{51.1} \\
    
    \bottomrule[0.95pt]
\end{tabular}
\vspace{-0.5mm}
\end{table}

\subsection{More Visual Instruction Tuning Experiments}
\label{sec:more_vlm}
% 放1% ratio的实验结果即可
In the main text, we present the performance of SEED on the visual instruction task under a $5\%$ selection ratio. Here, we further report its performance under a $1\%$ selection ratio in Table~\ref{tab:vlm2}.

\begin{table}[h]
\centering
\vspace{0mm}
\caption{\textbf{Performance of SEED on visual instruction tuning under 1\% selection ratio.}}
\vspace{0mm}
\footnotesize
\label{tab:vlm2}
\renewcommand{\arraystretch}{1.2}
\setlength{\tabcolsep}{0.7mm}

\begin{tabular}{p{1.6cm}p{.75cm}|ccccccc|c}
    \toprule[0.95pt]
    \multicolumn{2}{c|}{\textbf{Method}}
    & \scshape\textbf{Mathvista} 
    & \scshape\textbf{RealWorld} 
    & \scshape\textbf{MathVerse} 
    & \scshape\textbf{Hallusion} 
    & \begin{tabular}{c}
    \scshape\textbf{CharXiv}\\[-0.8mm]
    {\scriptsize (RQ/DQ)}
    \end{tabular}
    & \scshape\textbf{SimpleVQA} 
    & \scshape\textbf{MME}
    & \scshape\textbf{Avg.} \\
    
    \cmidrule(lr){1-2}
    \cmidrule(lr){3-9}
    \cmidrule(lr){10-10}
    
    \scshape\textbf{Base} & \texttt{\scriptsize{(0\%)}} & 72.4 & 71.2 & 41.2 & 56.7 & 41.2 / 80.7 & 47.9 & 2279.5 & 61.6 \\
    
    \rowcolor{mygray}
    \scshape\textbf{Full} & \texttt{\scriptsize{(100\%)}} & 78.4 & 73.7 & 51.8 & 55.9 & 50.8 / 75.8 & 47.5 & 2429.7 & 65.1  \\
    
    \midrule[0.6pt]
    
    \scshape\textbf{Random} &  & 71.1 & 72.3 & 44.1 & 53.5 & 45.5 / 74.5 & 47.3 & 2397.4 & 61.7 \\
    
    \scshape\textbf{LESS} \cite{xia2024less} 
    & \texttt{\scriptsize{ICML24}} & 73.9 & 69.9 & \textbf{52.1} & 55.0 & 46.2 / \textbf{79.4} & 48.8 & 2363.1 & 63.7 \\
    
    \scshape\textbf{Tailor} \cite{yu2025mastering} 
    & \texttt{\scriptsize{ICCV25}} & 73.5 & 72.0 & 48.5 & \textbf{55.6} & \textbf{50.2} / 74.5 & \textbf{49.5} & 2370.3 & 63.6 \\
    
    \rowcolor{mygray}
    \scshape\textbf{SEED} & & \textbf{74.1} & \textbf{74.0} & 51.5 & 55.1 & 48.1 / \textbf{79.4} & 48.0 & \textbf{2416.8} & \textbf{64.6} \\
    \bottomrule[0.95pt]
\end{tabular}
\vspace{-0.5mm}
\end{table}

\subsection{More Experiments with Small Proxies}
\label{sec:transfer_full}
As discussed in §\ref{sec:exp}, we assess the cross-family transferability of our data selection strategy using a broad model pool. While the main text highlights four representative cases, we provide the full results for all eight proxy-target pairs in Table~\ref{tab:full_transfer}. It demonstrates the robustness and scalability of SEED.
\begin{table}[h]
    \centering
	\caption{\textbf{Performance of SEED under proxy-driven selection settings.} Each group pairs a smaller proxy model with a larger target model, with progressively increasing scale gaps across heterogeneous model settings and varying data selection ratios. SEED-T denotes the use of the transfer strategy.}
    \begin{minipage}[t]{0.48\textwidth}
	\renewcommand\arraystretch{1.3}
	\setlength\tabcolsep{1.1mm}
	\centering
	\vspace{1mm}
	\label{tab:full_transfer}
	\footnotesize
	%\scriptsize
	%\normalsize
    \begin{tabular}{l|ccc|c}
        \toprule[0.95pt]
        \textbf{Method} & \scshape\textbf{TyDiQA} & \scshape\textbf{MMLU} & \scshape\textbf{BBH} & \scshape\textbf{Avg.} \\
        \midrule[0.6pt]
        \multicolumn{5}{l}{\scriptsize{Proxy Model: \raisebox{-0.25\height}{\includegraphics[height=1.05em]{fig/llama.png}}\hspace{0.2em}LLaMA2-\textbf{7B}} (Scale: \textbf{1.14}$\times$)} \\
        \scriptsize{\raisebox{-0.25\height}{\includegraphics[height=1.05em]{fig/llama.png}}\hspace{0.2em}LLaMA3-\textbf{8B}} & 47.3 & 63.3 & 62.2 & 57.6\\
        \scshape\textbf{Full} 100\% & 55.1 & 65.1 & 67.3 & \underline{62.5}\\
        \scshape\textbf{Random} 5\% & 51.6 & 64.0 & 63.1 & 59.6\\
        \rowcolor{mygray}
        \scshape\textbf{SEED}-T 5\% & 59.0 & 65.7 & 66.8 & \textbf{63.8}\\
        
        \midrule[0.6pt]
        \multicolumn{5}{l}{\scriptsize{Proxy Model: \raisebox{-0.25\height}{\includegraphics[height=1.05em]{fig/llama.png}}\hspace{0.2em}LLaMA2-\textbf{7B}} (Scale: \textbf{1.86}$\times$)} \\
        \scriptsize{\raisebox{-0.25\height}{\includegraphics[height=1.05em]{fig/llama.png}}\hspace{0.2em}LLaMA2-\textbf{13B}} & 38.7 & 54.0 & 46.5 & 46.4\\
        \scshape\textbf{Full} 100\% & 44.6 & 54.9 & 53.7 & \textbf{51.1}\\
        \scshape\textbf{Random} 5\% & 42.0 & 53.8 & 47.6 & 47.8\\
        \rowcolor{mygray}
        \scshape\textbf{SEED}-T 5\% & 47.3 & 55.7 & 50.1 & \underline{51.0}\\
        
        \midrule[0.6pt]
        \multicolumn{5}{l}{\scriptsize{Proxy Model: \raisebox{-0.25\height}{\includegraphics[height=1.3em]{fig/qwen.png}}\hspace{0.2em}Qwen3-\textbf{4B}} (Scale: \textbf{2.00}$\times$)} \\
        \scriptsize{\raisebox{-0.25\height}{\includegraphics[height=1.3em]{fig/qwen.png}}\hspace{0.2em}Qwen3-\textbf{8B}} & 36.5 & 72.7 & 78.3 & 62.5\\
        \scshape\textbf{Full} 100\% & 52.9 & 75.6 & 73.1 & 67.2\\
        \scshape\textbf{Random} 1\% & 54.0 & 72.9 & 78.6 & \underline{68.5}\\
        \rowcolor{mygray}
        \scshape\textbf{SEED}-T 1\% & 55.9 & 75.2 & 79.5 & \textbf{70.2}\\
        
        \midrule[0.6pt]
        \multicolumn{5}{l}{\scriptsize{Proxy Model: \raisebox{-0.25\height}{\includegraphics[height=1.3em]{fig/qwen.png}}\hspace{0.2em}Qwen3-\textbf{4B}} (Scale: \textbf{2.00}$\times$)} \\
        \scriptsize{\raisebox{-0.25\height}{\includegraphics[height=1.05em]{fig/llama.png}}\hspace{0.2em}LLaMA3-\textbf{8B}} & 47.3 & 63.3 & 62.2 & 57.6\\
        \scshape\textbf{Full} 100\% & 55.1 & 65.1 & 67.3 & \underline{62.5}\\
        \scshape\textbf{Random} 1\% & 52.3 & 65.3 & 66.3 & 61.3\\
        \rowcolor{mygray}
        \scshape\textbf{SEED}-T 1\% & 54.8 & 66.0 & 68.4 & \textbf{63.1}\\
        
        \bottomrule[0.95pt]
    \end{tabular}
    \end{minipage}
    \hspace{2mm}
    \begin{minipage}[t]{0.48\textwidth}
    \renewcommand\arraystretch{1.3}
	\setlength\tabcolsep{1.2mm}
	\centering
	%\begin{center}
	\vspace{1mm}
	\footnotesize
	%\scriptsize
	%\normalsize
    \begin{tabular}{l|ccc|c}
        \toprule[0.95pt]
        \textbf{Method} & \scshape\textbf{TyDiQA} & \scshape\textbf{MMLU} & \scshape\textbf{BBH} & \scshape\textbf{Avg.} \\
        \midrule[0.6pt]
        \multicolumn{5}{l}{\scriptsize{Proxy Model: \raisebox{-0.25\height}{\includegraphics[height=1.3em]{fig/qwen.png}}\hspace{0.2em}Qwen3-\textbf{1.7B}} (Scale: \textbf{4.12}$\times$)} \\
        \scriptsize{\raisebox{-0.25\height}{\includegraphics[height=1.05em]{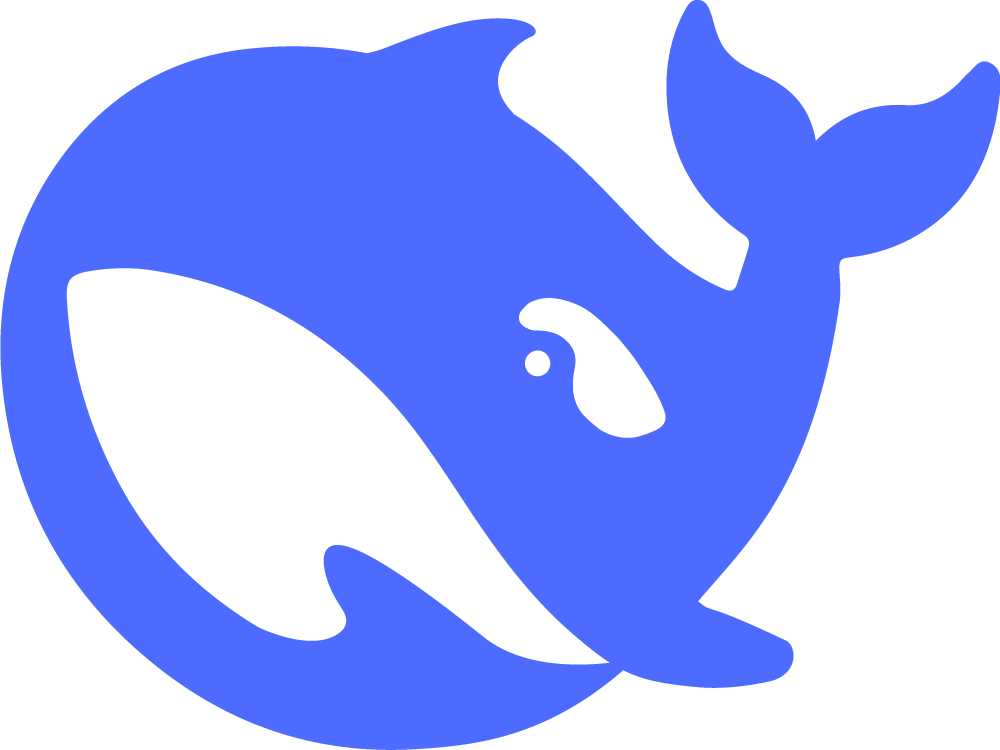}}\hspace{0.2em}Deepseek-\textbf{7B}} & 35.4 & 47.0 & 42.4 & 41.6\\
        \scshape\textbf{Full} 100\% & 40.9 & 47.5 & 44.6 & \underline{44.3}\\
        \scshape\textbf{Random} 5\% & 39.0 & 47.8 & 43.7 & 43.5 \\
        \rowcolor{mygray}
        \scshape\textbf{SEED}-T 5\% & 41.9 & 49.0 & 44.7 & \textbf{45.2}\\

        \midrule[0.6pt]
        \multicolumn{5}{l}{\scriptsize{Proxy Model: \raisebox{-0.25\height}{\includegraphics[height=1.05em]{fig/llama.png}}\hspace{0.2em}LLaMA2-\textbf{7B}} (Scale: \textbf{5.14}$\times$)} \\
        \scriptsize{\raisebox{-0.25\height}{\includegraphics[height=1.05em]{fig/bytedance.png}}\hspace{0.2em}Seed-OSS-\textbf{36B}} & 54.5 & 81.0 & 71.5 & 69.0 \\
        \scshape\textbf{Full} 100\% & 59.0 & 80.5 & 72.1 & 70.5 \\
        \scshape\textbf{Random} 5\% & 59.5 & 81.1 & 71.5 & \underline{70.7}\\
        \rowcolor{mygray}
        \scshape\textbf{SEED}-T 5\% & 60.5 & 83.3 & 72.4 & \textbf{72.1}\\
     
        \midrule[0.6pt]
        \multicolumn{5}{l}{\scriptsize{Proxy Model: \raisebox{-0.25\height}{\includegraphics[height=1.3em]{fig/qwen.png}}\hspace{0.2em}Qwen3-\textbf{1.7B}} (Scale: \textbf{8.24}$\times$)} \\
        \scriptsize{\raisebox{-0.25\height}{\includegraphics[height=1.05em]{fig/microsoft.png}}\hspace{0.2em}Phi-4-\textbf{14B}} & 49.4 & 80.3 & 69.0 & 66.2\\
        \scshape\textbf{Full} 100\% & 54.1 & 82.0 & 73.1 & \underline{69.7}\\
        \scshape\textbf{Random} 5\% & 53.7 & 81.1 & 71.8 & 68.9\\
        \rowcolor{mygray}
        \scshape\textbf{SEED}-T 5\% & 55.8 & 83.2 & 73.2 & \textbf{70.7}\\

        \midrule[0.6pt]
        \multicolumn{5}{l}{\scriptsize{Proxy Model: \raisebox{-0.25\height}{\includegraphics[height=1.05em]{fig/llama.png}}\hspace{0.2em}LLaMA2-\textbf{7B}} (Scale: \textbf{10.00}$\times$)} \\
        \scriptsize{\raisebox{-0.25\height}{\includegraphics[height=1.05em]{fig/llama.png}}\hspace{0.2em}LLaMA2-\textbf{70B}} & 41.3 & 68.2 & 65.4 & 58.3\\
        \scshape\textbf{Full} 100\% & 53.0 & 69.4 & 69.2 & \textbf{63.9} \\
        \scshape\textbf{Random} 1K & 46.8 & 68.3 & 66.0 & 60.4 \\
        \rowcolor{mygray}
        \scshape\textbf{SEED}-T 1K & 51.2 & 69.7 & 67.1 & \underline{62.7} \\
        
        \bottomrule[0.95pt]
    \end{tabular}
    \end{minipage}
    \vspace{-2mm}
\end{table}

\section{Sensitivity Analysis}
\label{sec:Sensitivity_Analysis}
We analyze the sensitivity of SEED to two hyperparameters in edge construction: the number of nearest neighbors $k$ and the local scaling factor $\alpha$. Both parameters govern the adaptive threshold $\tau_i$ in Eq.~\eqref{eq:adaptive_threshold}, and their appropriate configuration is critical to constructing a balanced conflict graph.

\begin{figure}[h]
    \centering
    \vspace{-1mm}
    \includegraphics[width=0.95\linewidth]{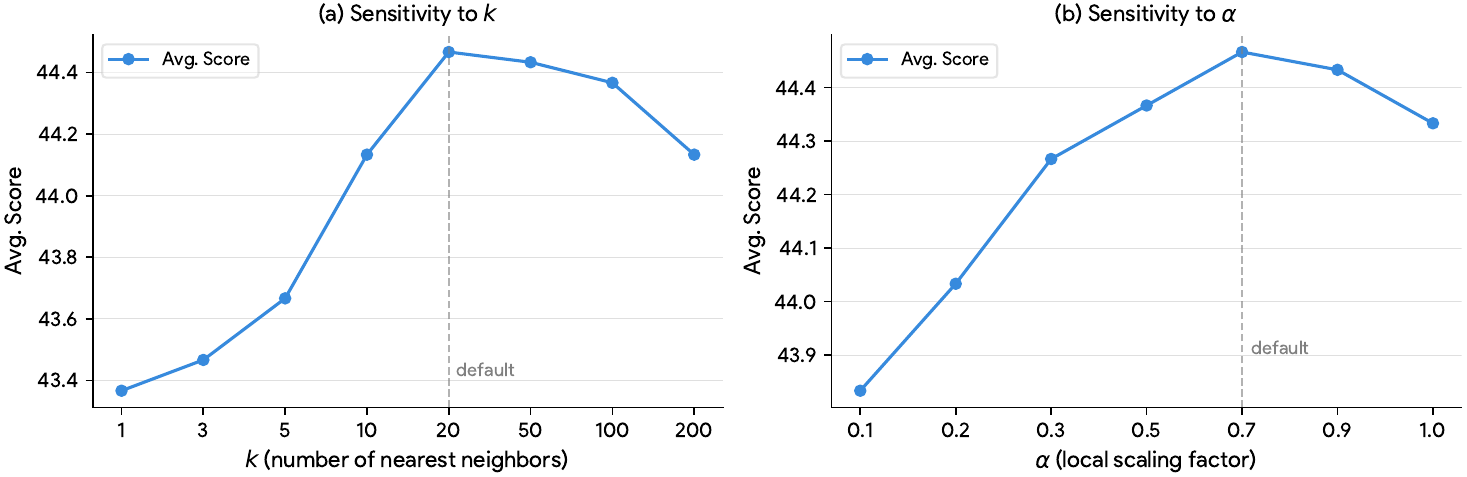}
    \caption{
        \textbf{Sensitivity analysis of SEED to $k$ and $\alpha$.} Results are reported on the LLaMA2-7B, measured as the average accuracy across TyDiQA, MMLU, and BBH under a $5\%$ data budget.
    }
    \label{fig:Sensitivity_Analysis}
    \vspace{-2mm}
\end{figure}

\textbf{Sensitivity to $k$.}
As shown in Figure~\ref{fig:Sensitivity_Analysis}(a), the average score peaks at $k=20$ and remains stable within $k \in [10, 50]$, with only marginal fluctuations.
Excessively small values ($k < 5$) degrade performance due to unstable local density estimates, while excessively large values ($k > 100$) over-smooth local structures, causing the adaptive threshold to degenerate toward a global one.
We set $k = 20$ as the default, balancing estimation stability and computational efficiency.

\textbf{Sensitivity to $\alpha$.}
As shown in Figure~\ref{fig:Sensitivity_Analysis}(b), the performance is consistent for $\alpha \in [0.5, 0.9]$ and peaks at $\alpha = 0.7$.
A smaller $\alpha$ causes $\tau_i$ to approach the global threshold $\tau$, effectively disabling local adaptation.
A larger $\alpha$ ($\alpha > 0.7$) aggressively suppresses edges in dense domains, leading to a slight performance drop.
We adopt $\alpha = 0.7$ as the default setting throughout all experiments.

Overall, the stability observed across hyperparameters confirms that SEED does not require careful per-dataset tuning, and the default configuration generalizes well across diverse tasks and models.

\section{Discussion}

\subsection{Limitations}
\label{sec:limitation}
In this work, we instantiate SEED using a static similarity graph construction and a standard MWIS-based optimization framework. While this design provides a principled and interpretable formulation for data selection, it does not fully exploit recent advances in dynamic graph construction \cite{mai2023dynamic}. Besides, SEED is primarily evaluated under instruction tuning and visual instruction tuning settings. Its effectiveness for other training paradigms, such as pre-training \cite{mckinzie2024mm1,zhang2025loss} or reinforcement learning \cite{pan2025timesearch,zhang2024unveiling} data selection, remains underexplored and is an interesting direction for future study.

\subsection{Societal Impacts}
\label{sec:impact}
SEED framework involves computing pairwise similarities and solving an MWIS problem over large-scale graphs, which introduces non-trivial computational overhead during data selection. However, it improves downstream training efficiency by selecting more informative and diverse subsets of data, thereby reducing redundancy in training data and lowering GPU costs and energy consumption, particularly in large-scale or repeated training scenarios. Therefore, despite the upfront cost of data selection, it contributes to more efficient and sustainable training pipelines in the long term.

%%%%%%%%%%%%%%%%%%%%%%%%%%%%%%%%%%%%%%%%%%%%%%%%%%%%%%%%%%%%

% \newpage
% \input{checklist.tex}

\end{document}